\documentclass[12pt]{article}
\usepackage[semicolon]{natbib}
\usepackage{amssymb}
\usepackage{amsmath}
\usepackage{hyperref}
\usepackage{multicol}
\usepackage{algorithm,algorithmic}
\usepackage{graphicx}
\usepackage{booktabs}
\usepackage{color}
\usepackage{float}
\usepackage{bbm,comment}
\usepackage{chngcntr}
\graphicspath{ {./images/} }
\usepackage[font=small,labelfont=bf, width=0.85\textwidth]{caption}
\newcommand{\bfx}{\mathbf{x}}

\newcommand{\bfw}{\mathbf{W}}

\newcommand{\bfa}{\mathbf{a}}
\newcommand{\bfb}{\mathbf{b}}

\title{Biologically Plausible Training Mechanisms for Self-Supervised Learning in Deep Networks}

\author{{\small\bf Mufeng Tang, Yibo Yang} \\  
{\small Department of Statistics} \\ {\small University of Chicago, Chicago, IL, USA} \\ {\small \{mufengt,yiboyang\}@uchicago.edu} \\
\and {\small\bf Yali Amit} \\ {\small Department of Statistics} \\ {\small University of Chicago, Chicago, IL, USA} \\ {\small amit@marx.uchicago.edu}}

\begin{document}
\date{}
\maketitle

\begin{abstract}

\vspace{5mm}\noindent We develop biologically plausible training mechanisms for self-supervised learning (SSL) in deep networks. Specifically, by biological plausible training we mean (i) All updates of weights are based on {\it current} activities of pre-synaptic units and current, or activity retrieved from short term memory of post synaptic units, including at the top-most error computing layer, (ii) Complex computations such as normalization, inner products and division are avoided (iii) Asymmetric connections between units, (iv)  Most learning is carried out in an unsupervised manner. SSL with a contrastive loss satisfies the third condition as it does not require labelled data and it introduces robustness to observed perturbations of objects, which occur naturally as objects or observer move in 3d and with variable lighting over time. We propose a contrastive hinge based loss whose error involves simple local computations satisfying (ii), as opposed to the standard contrastive losses employed in the literature, which do not lend themselves easily to implementation in a network architecture due to complex computations involving ratios and inner products. Furthermore we show that learning can be performed with one of two more plausible alternatives to backpropagation  that satisfy conditions (i) and (ii). The first is difference target propagation (DTP), which trains network parameters using target-based local losses and employs a Hebbian learning rule \citep{hebb1949organization}, thus overcoming the biologically implausible symmetric weight problem in backpropagation. The second is  layer-wise learning, where each layer is directly connected to a layer computing the loss error. The layers are either updated sequentially in a greedy fashion (GLL) or in random order (RLL), and each training stage involves a single hidden layer network. Backpropagation through one layer needed for each such network can either be altered with fixed random feedback (RF) weights as proposed in \citet{Lilli}, or using updated random feedback (URF) as in \citet{amit2019deep}. Both methods represent alternatives to the symmetric weight issue of backpropagation. By training convolutional neural networks (CNNs) with SSL and DTP, GLL or RLL, we find that our proposed framework achieves comparable performance to standard BP learning downstream linear classifier evaluation of the learned embeddings.

\end{abstract}

\newpage
\vspace{1.5in}

\newpage

\section{Introduction}

The rapid development of deep learning in recent years has raised extensive interest in applying artificial neural networks (ANNs) to the modelling of cortical computations. Multiple lines of research, including those using convolutional neural networks (CNNs) to model processing in the ventral visual stream \citep{yamins2016using, mcintosh2016deep}, and those using recurrent neural networks (RNNs) as models of generic cortical computation \citep{masse2019circuit, song2016training}, have suggested that ANNs are not only able to replicate behavioral activities (e.g. categorization) of biological systems, but are also capable of reproducing neuronal activities observed in vivo or in vitro. These observations, combined with the structural similarities (e.g. a hierarchy of layers and recurrent connections) between ANNs and cortical areas, make the use of deep learning a promising approach for modelling of neural computations.

Despite these successes, there are some fundamental problems confronting this approach. One issue with classification ANNs as a model for cortical learning is their reliance on massive amounts of labelled data and most ANN's employ backpropagation for learning. To be more specific, modern classification ANNs are trained to match their predictions to a set of target labels each associated with a training data point (e.g. an image), while biological systems, such as humans, usually learn without a large degree of supervision.
Most notably, backpropagation (BP), the learning rule for modern ANNs employing the chain rule for differentiation \citep{rumelhart1986learning}, is biologically implausible, as the same set of synaptic weights that have been used to compute the feedforward signals are also needed to compute the feedback error signals. Such a symmetric synaptic weight matrix does not exist in the brain \citep{amit2019deep, lillicrap2020backpropagation, zipser1993neurobiological}. 
In this paper by biologically plausible training we refer to an algorithm that satisfies the following conditions: (i) All updates of weights are based on {\it current} activities of pre-synaptic units and current, or activity retrieved from short term memory of post synaptic units, including at the top-most error computing layer, (ii) Complex computations such as normalization, inner products and division are avoided (iii) Asymmetric connections between units, (iv)  Most learning is carried out in an unsupervised manner.

In recent years, a number of solutions have been proposed to address  the symmetric connection issue \citep{lee2015difference, lillicrap2020backpropagation, amit2019deep, belilovskyGreedyLayerwiseLearning2019,akrout2019deep}. In parallel, in the deep learning community at large, there has been growing interest in self-supervised learning (SSL) where unlabeled data is used to create useful embeddings for downstream prediction tasks \citep{chen2020simple,he2020momentum,zbontar2021barlow,swav2020}. Rather than relying on labels as external teaching signals, SSL methods train neural networks with an objective function that attempts to maximize the agreement between two separate but related views of an input, each serving as the internal teaching signal for the other. Works such as \citet{chen2020simple} and \citet{he2020momentum} create the two views using random deformations on images, such as random crop and color jittering and other forms of deformation. Note that these deformations can be mapped into the motion of a real-world 3D object that provides the jittering effect in a natural environment, including changes in lighting. In parallel, the work in \citet{oord2018representation} and \citet{henaff2020data} explores this idea by creating pairs of views from neighboring patches in a larger image as surrogates for a gaze shift.
Since the agreement maximization between related views alone would cause the networks to produce a constant embedding regardless of the inputs (collapsing), different SSL methods employ different regularization terms to prevent collapsing, either using negative examples \citep{chen2020simple,he2020momentum,oord2018representation,henaff2020data}: images that are unrelated to the pair of related views; or using some constraints on the structure of the embedding, such as forcing the different embedding coordinates to be uncorrelated: \citet{zbontar2021barlow}, or well spread out in the embedding space \citep{swav2020}.

Self-supervised learning produces embeddings of the data with strong representation and robustness properties, in an unsupervised manner. Multiple lines of research have shown that embeddings from networks trained by SSL are comparable to those from networks trained by supervised methods in downstream tasks such as linear evaluation, namely training a linear classifier on labelled data using the learned embedding. These embeddings are also  robust to natural variation in the presentation of objects, performing particularly well in transfer learning \citep{chen2020simple,he2020momentum}. These properties of SSL, along with its unsupervised nature, make it an ideal model for cortical computations. Inheriting the patch-based method proposed by \citet{oord2018representation}, recent works \citep{lowe2019, illing2021} have used the patch-based SSL for biologically plausible learning by coupling them with localized learning rules and losses. In the meantime, \citet{zhuang2021unsupervised} used the deformation-based SSL methods proposed in \citet{chen2020simple} and \citet{he2020momentum} to train goal-driven ANNs that predicts neuronal activities with backpropagation. However, the deformation-based SSL methods per-se have not been investigated thoroughly as a computational model for learning in the brain.

\begin{figure}[t]
    \centering
    \noindent\makebox[\textwidth]{\includegraphics[width=0.8\paperwidth]{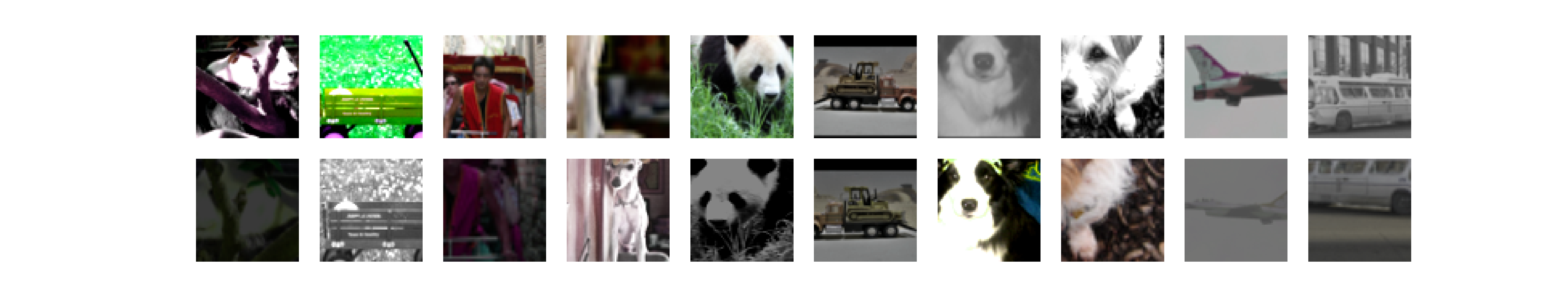}}
    \centering
    \caption{A more general way of getting shifting views of an object. Our deformation methods try to mimic the various possible ways of getting changed views of an object to form a positive pair. For example, the observation of a dog during two consecutive time steps may experience changing views caused by shortened distance and dimmed lighting. Effects of individual deformations can be found in Figure~\ref{fig:deform2}}
    \label{fig:deform1}
\end{figure}

In our work we further pursue this line of research initiated in \citet{lowe2019} and \citet{illing2021}, with a particular focus on deformation-based SSL. One contribution is to use a variety of global views of the whole object (see Figure~\ref{fig:deform1}) with the global deformation framework by \citet{chen2020simple}, which is a {\bf more general way of getting view changes} than the vertical patch movements in \citet{lowe2019} and \citet{illing2021} that typically focus on only small parts of an object. This appears to be more consistent with the real time input obtained when observing real world 3D objects. We then propose a self-supervised loss function with {\bf spatial locality} of computations, called  {\it contrastive hinge loss}, which only requires the simplest form of Hebbian updates using the outputs of the pre- and post-synaptic units \citep{hebb1949organization}. This loss avoids  computations such as normalization, inner product and divisions found in the losses used in \citet{chen2020simple}, \citet{oord2018representation} and \citet{he2020momentum}, and requires no global dendritic inputs employed in the loss by \citet{illing2021}. We push our method further towards {\bf temporal locality} by considering the first image of a positive pair in SSL as stored in `short-term' memory, which is then compared to the second positive pair, or a negative example. Thus the loss is only computed after the second image is observed, and the gradient can only be be affected by the embedding of the second image. We thus make sure that the gradient of the loss does not pass through the first observed image to ensure temporal locality of our approach.

We then combine our SSL framework with several biologically plausible learning methods proposed for supervised learning in previous works, but with our own novel modifications. We first explore {\bf Randomized Layer-wise Learning with Random Feedback}. In layer-wise learning, each layer has a direct connection to the layer computing the self-supervised loss, and updates a single hidden layer with input given by the output of the previous layer. The sequential form of layer-wise learning  has been studied in a number of papers  \citep{hinton2006fast, bengio2007greedy,huang2018learning,belilovskyGreedyLayerwiseLearning2019,jaderbergDecoupledNeuralInterfaces2017,noklandTrainingNeuralNetworks2019}, which nevertheless requires rigid timing of the updates of each layer. This would seem unlikely in the biological setting. We thus propose the Randomized Layer-wise Learning (RLL), which randomizes the layers being updated at each step, for a more plausible mechanism. Since training the single hidden layer with BP introduces the weight symmetry problem, we also explore the possibility of using the random feedback (RF) approach of \citet{Lilli}, which replaces the symmetric feedback weight in BP with a random one. This method works well with shallow networks (in this case one hidden layer) but its performance deteriorates with the depth of the network \citep{bartunov2018assessing}. It is thus quite suitable for layer-wise learning. \citet{amit2019deep} proposed updated random feedback (URF), which uses the same updates to the forward and backward connections starting at random initial conditions as opposed to imposing strict symmetry. This approach works
as well as BP on shallow networks. We also investigate {\bf Difference Target Propagation with pooling layers}. Difference target propagation (DTP) \citep{lee2015difference} updates the weights in a network by minimizing a set of local losses, computed as the difference between the bottom-up forward activities and the top-down targets, both propagated through a set of non-linear functions. The localized losses in DTP then yield a Hebbian learning rule for the connectivity weights. Previous works \citep{bartunov2018assessing} suggested that pooling layers in CNNs are not compatible with this learning rule, and used strided convolutional layers to perform down-sampling of input data. We nevertheless find that pooling layers can in fact be incorporated into networks trained by DTP. With our SSL method, we show that DTP-pooling achieves comparable results to BP on CIFAR10 \citep{krizhevsky2009learning}, whereas previous works on supervised DTP \citep{bartunov2018assessing} have presented a performance gap between DTP and BP on this dataset.

Finally, we use a simple experiment to demonstrate the {\bf robustness of our embedding to object variability} in downstream tasks, where the labeled training data contain a more limited range of variability. Although traditionally this is resolved by using data augmentation on the labeled training data, this shows that robustness can be achieved a-priori with unlabeled data.
 
This paper is organized as follows. In section 2 we discuss related works in SSL  and in biologically plausible learning rules. In section 3 we describe the technical details of our proposed framework, including our biologically plausible SSL method, DTP-pooling, and RLL with RF. In section 4 we present the experimental results on three different datasets: STL10 \citep{stl10}, CIFAR100/CIFAR10 \citep{krizhevsky2009learning} and EMNIST/MNIST \citep{cohen2017emnist, lecun-mnisthandwrittendigit-2010}, including our proposed framework's performance in linear evaluation and transfer learning tasks. We conclude with a discussion section.

\section{Related Work}

{\bf Biologically Plausible Learning Rules.} The biological implausibility of backprop was mentioned in \citet{zipser1993neurobiological} and they first suggested to separate the feedforward weights of ANNs from the feedback weights. More recently, \citet{Lilli} proposed a biologically plausible learning rule called `feedback alignment' (FA), which decouples the feedforward and feedback weights by fixing the feedback weights at random values, we rename this `random feedback' (RF). However, this method does not scale well to deeper networks and more challenging datasets \citep{amit2019deep}. Several modifications have been proposed to improve the performance of FA. \citet{liao2016important} found that using Batch Normalization (BN) could improve the performance of FA, but it is unclear how BN can be employed by biological neural circuits. 

Another track of research on the improvement of FA focuses on finding a learning rule for the feedback weights (rather than fixing them).  \citet{amit2019deep} proposed to train the randomly initialized feedback weights using the same updates as those for the  feedforward weights, and found a significant improvement of error rates in deeper networks, hence the name `updated random feedback' (URF). This work also paid particular attention to the topmost layer. By modifying the loss for supervised learning, the learning rule at the topmost layer in this method yields a Hebbian update and is thus more biologically plausible than the softmax loss.  In our work we follow this idea and propose a more biologically plausible loss for SSL. Similar to \citet{amit2019deep}, \citet{akrout2019deep} also proposed a learning rule for the feedback weights that will force the feedback weights to converge to the feedforward weights, resulting in a convergence to backprop. In \citet{illing2021} both RF and and URF are explored as methods to update each trained layer.

An alternative modification of deep learning yielding more biologically plausible update rules is DTP \citep{lee2015difference}. Localized losses are introduced in all layers, such that the weight updates are purely local and independent of the outgoing weights. The backward computation in DTP propagates the `targets' top-down and uses a set of backward weights learned through layer-wise autoencoders. This structure on the basis of layer-wise losses connects DTP to predictive coding \citep{rao1999predictive, whittington2017approximation}, which have formed a well-established computational model for brain areas such as visual cortex. However, these latter works on predictive coding still suffer from the symmetric weight problem, which is avoided in DTP. \citet{bartunov2018assessing} further investigated DTP with CNNs and more challenging datasets such as CIFAR10, and found a significant performance gap between backprop and DTP in supervised visual tasks such as classification. DTP has also been applied to RNNs \citep{manchev2020target}, and variants of DTP, such as that in \citet{ororbia2020large}, have been proposed recently to address the slow training issue due to the layer-wise autoencoders in the original algorithm. \citet{MeulemansTP2020} developed a theoretical analysis of Target Propagation, showing that this biologically more plausible learning rule in fact approximates Gauss-Newton optimization and is thus significantly different from BP.

In end-to-end learning, all layers in the network, after passing the input signal forward to the next layer, must wait for the signal to feed-forward through the rest of the network and the error signal to propagate back from the last layer. No updates can be done during this period. This constraint is referred to as the locking problem by  \citet{jaderbergDecoupledNeuralInterfaces2017}. Moreover end-to-end learning requires some mechanism of passing information sequentially through multiple layers, whether using backpropagation or target propagation. Layer-wise learning is an alternative to end-to-end learning that tackles both problems. Greedy unsupervised layer-wise learning was first proposed to improve the initialization of deep supervised neural networks \citep{hinton2006fast, bengio2007greedy}. \citet{huang2018learning} used the layer-wise method to train residual blocks in ResNet sequentially, then refined the network with the standard end-to-end training. \citet{belilovskyGreedyLayerwiseLearning2019} studied the progressive separability of layer-wise trained supervised neural networks and demonstrated Greedy Layer-wise Learning (GLL) can scale to large-scale datasets like ImageNet. Other attempts at supervised layer-wise learning involve a synthetic gradient \citep{jaderbergDecoupledNeuralInterfaces2017} and a layer-wise loss that combines local classifier and similarity matching loss \citep{noklandTrainingNeuralNetworks2019}.

\vspace{4mm}\noindent 
{\bf Self-supervised Learning.} The idea of SSL has been proposed in \citet{Becker1992SelforganizingNN}. The authors of this work used a self-supervised objective that maximizes the agreement between the representations of two related views of an input, subject to how much they both vary as the input is varied. More recent work in the context of deep learning can be found in the Contrastive Predictive Coding (CPC) framework by \citet{oord2018representation} where nearby patches from the larger images are used as positive pairs. The methods in \citet{chen2020simple} and \citet{he2020momentum} use standard data augmentation and deformation methods such as cropping, resizing, color jittering etc. to create the positive pairs. These methods introduced the concept of contrasts to prevent collapse, where the self-supervised objective includes a term that maximizes the agreement between related representations (`positives'), as well as a term to minimize the agreement between unrelated representations (`negatives'), thus preventing  the networks from producing a constant output regardless of the inputs. Optimizing this objective will create contrasts between positives and negatives, hence the term  {\it contrastive learning}.

The acquisition of positives is similar across deformation-based SSL methods, using random perturbations of the input. The selection of negatives is what differentiates them. The $\textsc{SimCLR}$ framework, proposed by \citet{chen2020simple}, uses all other images within a mini-batch as negatives of an image. In this way, the batch size is associated with the number of negatives and usually needs to be large enough to provide sufficient negatives. To decouple these two hyperparameters, the $\textsc{MoCo}$ framework, by \citet{he2020momentum}, uses a dynamic queue of negatives, with its length decoupled from the batch size. During training, every new mini-batch is enqueued and the oldest mini-batch is dequeued, which provides a larger sample of negatives from the continuous space of images. 

More recently, a few new methods in SSL have been proposed to eliminate the need for negatives (therefore no contrast), including $\textsc{BYOL}$ \citep{grill2020bootstrap}, $\textsc{SimSiam}$ \citep{chen2020exploring} and Barlow Twins \citep{zbontar2021barlow}. $\textsc{BYOL}$ and $\textsc{SimSiam}$ both used a top-layer linear predictor and a gradient block to asymmetrize the two networks for the pair of positives, and the performance of their idea was proved to be comparable to the contrastive methods. Recent theoretical works have also shed light on why the linear predictor and gradient block help prevent collapse \citep{ganguli2021}. Barlow Twins uses a symmetric architecture, and their objective function enforces the cross-correlation matrix between the positives to be as close to an identity matrix as possible. Its objective has a biological interpretation closely related to the redundancy reduction principle \citep{barlow1961possible}, which has been used to describe how the cortical areas process sensory inputs. We do not experiment with Barlow Twins in this paper, as we are still exploring biologically plausible ways to implement this loss.

\vspace{4mm}\noindent 
{\bf Relationship of SSL to biology}
Of interest is the work in \citet{zhuang2021unsupervised} which uses SSL to train a model for sensory processing in the ventral visual cortical areas. Using the techniques developed by \citet{yamins2016using} and \citet{schrimpf2018brain}, they measured the correlation between recordings of neuronal activities from the ventral visual stream and activities of CNNs trained by the deformation-based $\textsc{SimCLR}$, given the same visual inputs and tasks. The CNNs were found to achieve highly accurate neuronal activity predictions in multiple ventral visual cortical areas, even more so than CNNs trained with supervised learning. However, the CNN models in their experiments are not trained as computational models for learning in cortical areas, and thus are trained with BP and the $\textsc{SimCLR}$ loss, which would be difficult to compute using neural circuits (see section \ref{SSL}). In our work we suggest a model for learning, with biologically plausible learning rules and an alternative simpler loss that could be implemented with simple neural circuits.

The work in \citet{illing2021} adapts some of the more biologically plausible learning rules mentioned above to the SSL context. Inspired by \citet{he2020momentum}, they employ pairs of nearby image patches (small 16x16 sub-images) as positive pairs, motivated by the effect of small eye movements. Negative pairs are created from patches in other images. They also employ a layer-wise learning mechanism, with a loss that computes the inner product of the embedding of the first image of the pair, to a linear transformation of the embedding of the second image. However, the learning rule derived from this loss requires global dendritic inputs from a group of neurons, rather than local and individual pre- and post-synaptic activities. 
The embedding and the linear transformation are all trained parameters. They note that there is no need
to use a large number of negative examples for each positive example, thus making training more realistic in that there is no need to retain a buffer of multiple negative examples. They also explore updating the different layers synchronously, and use more biologically plausible update rules for updating each layer similar to those used in \citet{amit2019deep}.

\section{Methods}

\noindent In this section we first describe the SSL framework, with a particular focus on how we push it towards a biologically more plausible learning model by modifying its architecture to achieve temporal locality, and introducing a contrastive hinge loss with spatial locality of computations. We then describe how we use RF, layer-wise learning and DTP to address the symmetric weight problem in BP and our modifications to improve their performance in deep convolutional networks.

\subsection{SSL with global views and local computations}\label{SSL}

\begin{figure}[t]
    \centering
    \noindent\makebox[\textwidth]{\includegraphics[width=0.75\paperwidth]{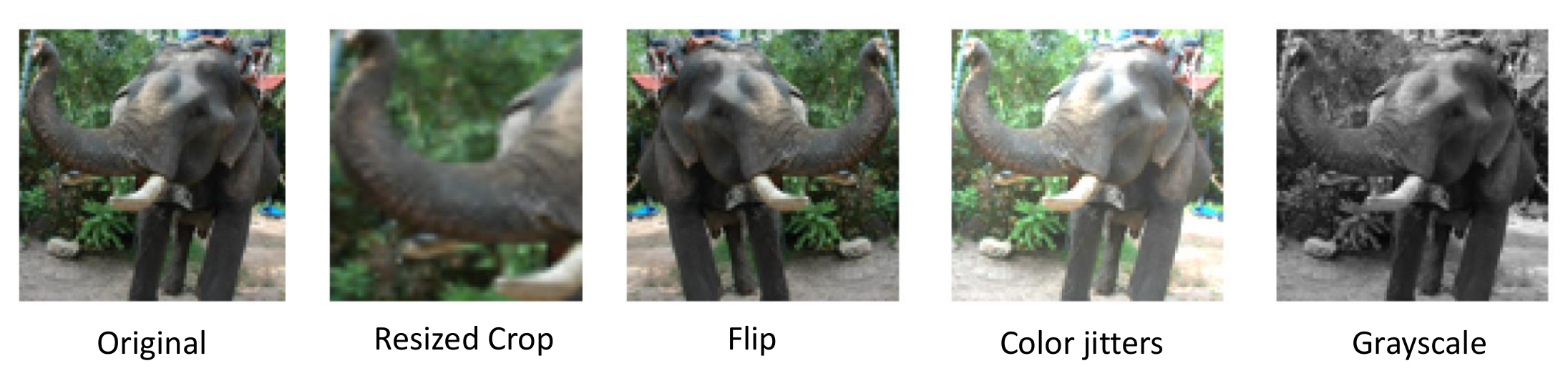}}
    \centering
    \caption{Individual deformations used in our SSL method. Each of them is applied randomly to each image in a batch, creating a variety of views.}
    \label{fig:deform2}
\end{figure}

\begin{figure}[t]
    \centering
    \noindent\makebox[\textwidth]{\includegraphics[width=0.75\paperwidth]{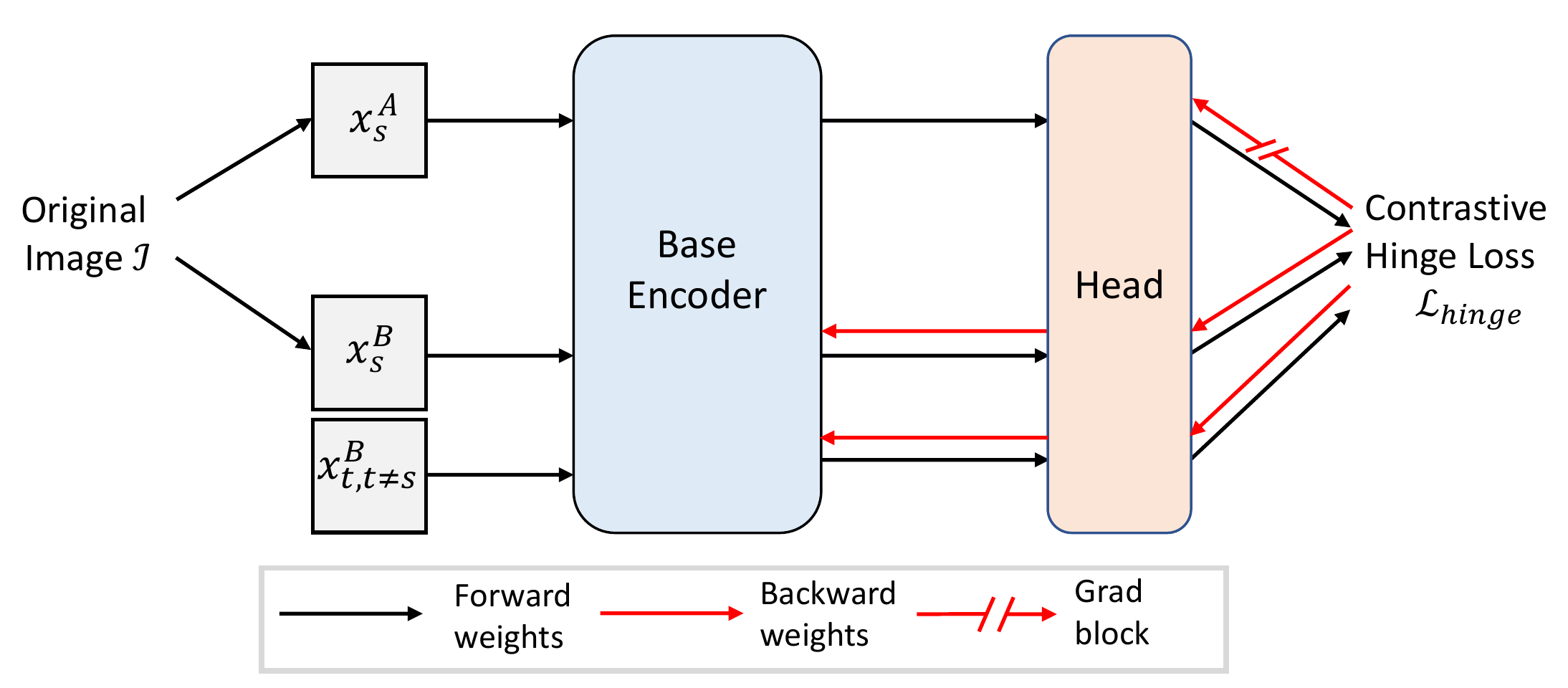}}
    \centering
    \caption{Overview of our proposed self-supervised model with contrastive hinge loss. The gradient flow in $\bfx_s^A$ is blocked. For demonstration purpose only one negative $\bfx_t^B,t\neq s$ is shown.}
    \label{fig:ssl}
\end{figure}

Given an input image $\mathcal{I}$, deformation-based SSL first applies two different sets of random deformations to it to create a pair of positive samples. We denote this pair as ${\bf x}^A_{s}$ and ${\bf x}^B_{s}$, indicating that they are the input into the 0-th layer of our networks (corresponding to the retina). $s\in\{1,...,n\}$ indicates their batch index within a batch of size $n$. Specifically, we apply 1) random resized crop, 2) random flip, 3) random color jittering (including changes in contrasts, brightness, saturation and hue), and 4) random changes to grayscale images, resulting in a variety if global views of a whole object. This whole range of deformations can be mapped into the observation of the real-time motion of real-world 3D objects (see Figure~\ref{fig:deform2}), making our framework more consistent with how animals perceive `positive pairs'. Notice that in \citet{lowe2019} and \citet{illing2021}, consecutive patches (sub-images) of fixed sizes following a vertical order were used to create positive pairs of views, to model the gaze shift during observation of an object, whereas in our work, the random resized crop provides shifting views of the object towards random angles with random sizes of the field of views. \citet{lowe2019} and \citet{illing2021} also transformed all images to grayscale, while the color jittering in our deformations reflects the change of real-world lighting conditions. 

We denote the set of unrelated negative samples as ${\bf x}^B_{t}, t \neq s$. We follow the scheme developed by \citet{chen2020simple} and select our negatives from the current batch of images fed into the network. However, instead of using the whole batch of unrelated images as in \citet{chen2020simple}, we only select $T$ negative samples from the batch indicating a small number of closest observations to the original image along the temporal dimension. This reflects our consideration that the memory of the original image can only be stored for a relatively short period of time, using the models for working memory proposed in works such as \citet{tsodyks1} and \citet{tsodyks2}.

The positive pair and the negative samples are then passed into the same encoder network with $L$ layers, and a single-layer projection head to produce their embeddings ${\bf x}^A_{L,s}$, ${\bf x}^B_{L,s}$ and ${\bf x}^B_{L,t}, t \ne s$. The whole network is then trained to minimize the distance between ${\bf x}^A_{L,s}$ and ${\bf x}^B_{L,s}$, i.e. to maximize the agreement between the two related views, and to maximize the distance between ${\bf x}^A_{L,s}$ and ${\bf x}^B_{L,t}$'s in order to prevent collapsing. \citet{chen2020simple} achieved this objective using a multinomial logistic loss function and we refer to it as the $\textsc{SimCLR}$ loss in our work:

\begin{equation}\label{simclr}
\begin{aligned}
\mathcal{L}_{\textsc{SimCLR}}=-\sum_{s=1}^n\log\frac{\exp(\text{sim}({\bf x}^A_{L,s},{\bf x}^B_{L,s})/\tau)}{\sum_{t\neq s}^n\exp(\text{sim}({\bf x}^A_{L,s},{\bf x}^A_{L,t})/\tau) + \sum_{t=1}^n\exp(\text{sim}({\bf x}^A_{L,s},{\bf x}^B_{L,t})/\tau)} \\
-\sum_{s=1}^n\log\frac{\exp(\text{sim}({\bf x}^B_{L,s},{\bf x}^A_{L,s})/\tau)}{\sum_{t\neq s}^n\exp(\text{sim}({\bf x}^B_{L,s},{\bf x}^B_{L,t})/\tau) + \sum_{t=1}^n\exp(\text{sim}({\bf x}^B_{L,s},{\bf x}^A_{L,t})/\tau)}
\end{aligned}
\end{equation}

\noindent where $\text{sim}(\bfa,\bfb)=\bfa\cdot\bfb / \Vert\bfa\Vert_2 \Vert\bfb\Vert_2$ and $\tau$ denotes a temperature hyperparameter. 

We use the $A,B$ notation, which is different from that of \citet{chen2020simple}, to emphasize that in our proposed loss (see below), for each anchor image $x^A$ we only use negatives from the `B' list, whereas in $\textsc{SimCLR}$ negative images come from both the A-list and the B-list so that it is entirely symmetric in terms of the two branches. 
To compute the update of an individual weight $W_{L,ij}$ in the topmost layer, connecting the pre-synaptic neuron $j$ and post-synaptic neuron $i$, it is necessary to compute complex ratios and inner products related to the output of the other neurons in this layer, making the learning rule for the topmost layer non-local.  Moreover, to make sure the similarity measurements are at the same scale across all data points, all of the embeddings have to be normalized to unit norm. Although some evidence has been found for this type of computation in cortex (see \citet{carandini2012normalization}), the instances are usually at lower levels of sensory input and involve local computations with neurons of similar types of responses. In this case we require normalization over a set of neurons with very diverse responses to very high level functions of the input. Taking into account all these issues motivates the introduction of a simpler learning rule that avoids such operations, which, as we will see, does not degrade performance.

We propose an alternative contrastive loss inspired by the biologically plausible supervised loss in \citet{amit2019deep}, which requires only local activities of the neurons in the top layer, and does not require normalization, thus lending itself to very simple network implementations. We call it the {\it contrastive hinge loss}. For a batch of size $n$ with embedding $\left(\bfx^A_{L,1},\dots,\bfx^A_{L,n}\right)$ and $\left(\bfx^B_{L,1},\dots,\bfx^B_{L,n}\right)$, the loss is:

\begin{equation}
    \mathcal{L}_{\textnormal{hinge}}
    =\sum_{s=1}^n \left[ [\Vert \bfx^A_{L,s} - \bfx^B_{L,s} \Vert_1 - m_1]_+
    +\sum_{t \in {\cal N}_s}[m_2 - \Vert \bfx^A_{L,s} -  \bfx^B_{L,t} \Vert_1 ]_+ \right],
\end{equation}
where $s \notin {\cal N}_s$ is a subset of the $B$ batch. The set ${\cal N}_s$ could contain all of the batch or
at the other extreme just one negative example.
Only positives with a distance greater than a margin ($m_1$), and negatives with a distance smaller than a margin ($m_2$) will be selected, making sure that optimization only depends on `problematic' examples.  Notice that in the topmost layer the forward computation is $\bfx_L = \bfw_L \bfx_{L-1}$, and thus during learning the gradient descent update of $W_{L,ij}$ i.e. the weight connecting the pre-synaptic neuron $j$ to post-synaptic neuron $i$, due to a single anchor input $\bfx^A_{L,s}$ is as follows (omitting the $s$ subscript):

\begin{equation}\label{eq:update}
    \Delta W_{L,ij} \propto -\left(\delta^A_{L,i} x^A_{L-1,j}+\delta^B_{L,i}x^B_{L-1,j} + \sum_{t \in {\cal N}_s} \delta^B_{t,L,i} x^B_{t,L-1,j}\right)
\end{equation}
where the error signal of this loss is:
\begin{equation}\label{eq27}
    \delta^A_{L,i} = \mathbbm{1}_{\Vert \bfx^A_{L} - \bfx^B_{L} \Vert_1>m_1}sgn(x^A_{L,i} - x^B_{L,i}) - \sum_{t \in {\cal N}_s}\mathbbm{1}_{\Vert \bfx^A_{L} - \bfx^{B}_{t,L} \Vert_1 < m_2}sgn(x^A_{L,i} - x^{B}_{t,L,i})
\end{equation}
\begin{align}
\delta^B_{L,i}= &-\mathbbm{1}_{\Vert \bfx^A_{L} - \bfx^B_{L} \Vert_1>m_1}sgn(x^A_{L,i}-x^B_{L,i} )  \\
\delta^B_{t,L,i} = &\mathbbm{1}_{\Vert \bfx^A_{L} - \bfx^{B}_{t,L} \Vert_1 < m_2}sgn( x^A_{L,i}-x^{B}_{t,L,i} )
\end{align}

In the case of gradient blocking, we block the gradient through $\bfx^A_{L}$, hence the weight update in equation~\eqref{eq:update} becomes
\begin{equation}
    \Delta W_{L,ij} \propto -\left(\delta^B_{L,i}x^B_{L-1,j} + \sum_{t \in {\cal N}_s} \delta^B_{t,L,i} x^B_{t,L-1,j}\right)
\end{equation}

During learning, this error signal will only depend on activities local to the $i$th neuron in the topmost layer. The only information needed from other neurons in the output is in filtering out the `easy' negatives involving L1 distances between activities. 

In the extreme case where ${\cal N}_s$ contains one negative this update can be viewed as follows. The anchor image
${\bfx^A}$ is shown and the embedding ${\bfx^A_{L}}$ is retained in short term memory, as modeled for example in \citet{tsodyks1} and \citet{tsodyks2}.
Then $\delta^{B}_{L,i}$ is computed and the weight $W_{L,ij}$ updated with $\delta^{B}_{L,i} {x^B_{L,j}}$. The anchor image is still kept in short term memory
and once a saccade has occurred to some other object $\delta^{B}_{t,L,i}$ is computed and the same weight $W_{L,ij}$ is updated with $\delta^{B}_{t,L,i} {x^B_{L,j}}$.
If several negatives are required $x^A_{L,i}$ needs to be kept in short term memory for a longer period. Note that the
memory trace of ${x^A_{L,i}}$ needs to be stored in another unit, as the actual $i,L$ unit is activated with
$\delta^{B}_{L,i}$ or $\delta^{B}_{t,L,i}$, in order to perform the Hebbian update $\delta^{B}_{L,i} {x^B_{L,j}}$ or $\delta^{B}_{t,L,i} {x^B_{t,L,j}}$.
If the block gradient is not implemented it would be necessary to retain the input ${\bfx^{A}_{L-1}}$ in some short
term memory as well and retrieve it to the units in the $L-1$ layer for the update $\delta^A_{L,i}x^A_{L-1,j}$.
It is in this sense that we view the blocked gradient implementation as yielding greater temporal locality for
the learning rule, in the context of small numbers of negatives.

The computation of the error signals $\delta^{B}_{L,i}$ requires the computation of the difference between
the activity $x^{A}_{L,i}$ stored in short term memory and the activity $x^{B}_{L,i}$. 
One can imagine the computation of the thresholds in the error signals  with rectified units and a combination of excitatory and inhibitory neurons (see Appendix \ref{threshold}). Moreover no normalization is needed.  Figure~\ref{fig:ssl} shows the general structure of our proposed self-supervised model.

\subsection{Weight Symmetry and Alternatives to BP}
For a multilayer network with layers $1,2,\dots, L$, we denote the activation value of the $l$th layer as $\mathbf{x}_l \in \mathbb{R}^{n_l}$. We denote the feedforward weight from the $(l-1)$th to the $l$th layer as $\mathbf{W}_l \in \mathbb{R}^{n_l\times n_{l-1}}$. The forward computation in this layer can be written as: 

\begin{equation} \label{eq2}
    \mathbf{x}_l = \sigma(\mathbf{h}_l), \quad \mathbf{h}_l = \mathbf{W}_l \mathbf{x}_{l-1}
\end{equation}

\noindent where $\sigma$ is the element-wise non-linearity. In the top layer, where no non-linearity will be applied, the forward computation will be $\mathbf{x}_L = \mathbf{W}_L \mathbf{x}_{L-1}$ and a global loss $\mathcal{L}(\mathbf{x}_L,\mathbf{y})$ will be computed based on the true labels vector $\mathbf{y}$ and final layer activities $\mathbf{x}_L$ in supervised learning.

The difficulty in imagining a biological implementation of BP has been discussed extensively \citep{zipser1993neurobiological,lillicrap2020backpropagation}  and boils down to the need for symmetric synaptic connections between neurons in order to propagate the error backwards through the layers. Following the feed-forward pass above, in BP the update of the weight matrix $\mathbf{W}_l$ is $\Delta \mathbf{W}_l = \frac{\partial \mathcal{L}}{\partial\mathbf{h}_l}\mathbf{x}_{l-1}^T = \delta_l\mathbf{x}_{l-1}^T$, where

\begin{equation} \label{eq:sym-weight}
    \delta_l = \frac{\partial\mathcal{L}}{\partial\mathbf{x}_l}\frac{\partial\mathbf{x}_l}{\partial\mathbf{h}_l} = \sigma^{\prime}(\mathbf{h}_l)\bfw_{l+1}^T\delta_{l+1}
\end{equation}

\noindent and $\sigma^{\prime}(\mathbf{h}_l)$ is a diagonal matrix with the $i$-th diagonal element being $\sigma^{\prime}(h_{l,i})$. The need for the outgoing weight $\bfw_{l+1}$ when computing the update of the weight matrix introduces the symmetric weight problem.

\vspace{4mm} \noindent
{\bf Random Feedback (RF).} Proposed by \citet{Lilli}, this learning rule tackles the symmetric weight problem by simply replacing the weight $\bfw_{l+1}^T$ in \eqref{eq:sym-weight} with a fixed random matrix $\mathbf{B}_{l+1}$, i.e. the error signal in RF is computed as:

\begin{equation} \label{eq:FAerrorSignal}
    \delta_{l}=\sigma^{\prime}(\mathbf{h}_l) \mathbf{B}_{l+1} \delta_{l+1}
\end{equation}

Although proven to be comparable to BP in shallow networks, the performance of RF degrades as network depth increases, due to the fact that the alignment between $\bfw_{l+1}$ and $\mathbf{B}_{l+1}$ weakens \citep{amit2019deep}. We thus combine RF with layer-wise learning, where the RF error signals \eqref{eq:FAerrorSignal} are always computed in a shallow, 1-layer network, to tackle this issue with depth.

\begin{figure}[t]
    \centering
    \noindent\makebox[\textwidth]{\includegraphics[width=0.6\paperwidth]{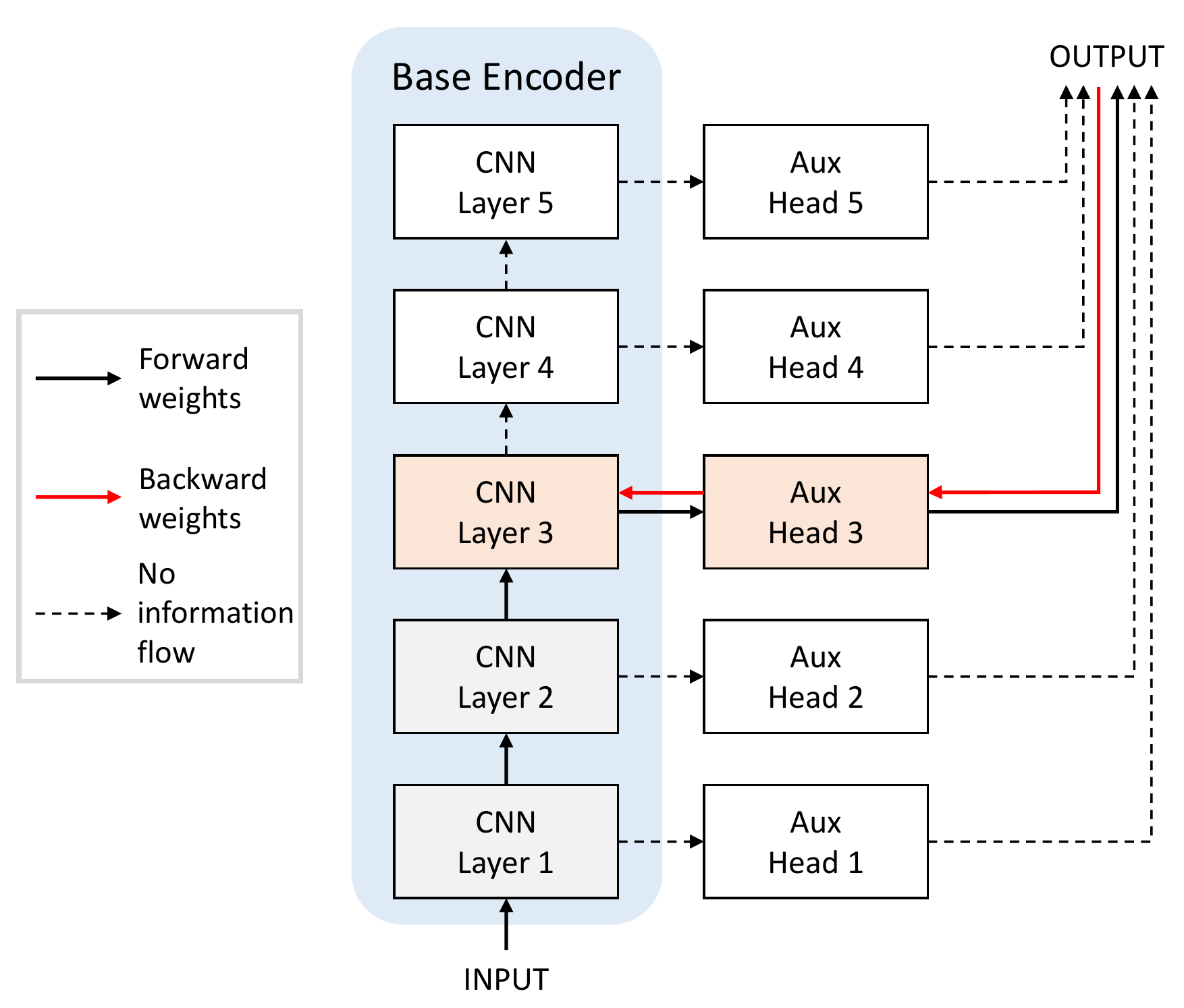}}
    \centering
    \caption{Training the base encoder in Figure \ref{fig:ssl} with layer-wise learning. When training layer 3, for example, only the shaded orange and grey blocks are included in the network, and only the orange blocks (layer 3 of the network and the auxiliary head 3) are updated. For Greedy Layer-wise Learning (GLL), the layers are trained till convergence sequentially from down to top. For Randomized Layer-wise Learning (RLL), a layer is randomly selected to train for each batch of training data. The feedback weights (red arrows) from head to encoder can be random in our proposal.}
    \label{fig:layerwise-diagram}
\end{figure}

\vspace{4mm} \noindent
{\bf Updated Random Feedback (URF).}
In \citet{amit2019deep}  a more plausible version of BP is proposed, where the random feedback weights $\mathbf{B}_{l}$, which are initialized differently from the feedforward weights, are updated
with the same increment as the feedforward weights, as defined in \eqref{eq:sym-weight}. The updated random feedback (URF) method yields results very close to BP, essentially indistinguishable for shallow networks. We compare the performance of RF with the one layer BP/URF in the context of layer-wise training.

\vspace{4mm}\noindent {\bf Greedy and Randomized Layer-wise Learning.} We adapted the supervised Greedy Layer-wise Learning (GLL)  method proposed in \citep{belilovskyGreedyLayerwiseLearning2019} to self-supervised learning by training convolutional layers sequentially with auxiliary heads and self-supervised loss, as shown in Figure~\ref{fig:layerwise-diagram}. The base encoders are trained layer by layer. For each layer, the new training layer is added on top of the previous architecture, and only parameters in the new training layer and its auxiliary head are updated. Thus at each step we are training a network with one hidden layer. 
Then, we replace the auxiliary heads with a linear classifier layer and only updates these weights with supervised learning to measure the representational power of the embedding. The pretrained network with self-supervised learning provides the encoder input to the classifier and is not updated. 

GLL tackles the locking problem as it does not require back-propagating through the whole network to get the full gradients. It does not require storage  of activations in intermediate layers. In addition to GLL, where we sequentially train layers with auxiliary heads, we also explore what we called Randomized Layer-wise Learning (RLL), where a random layer in the architecture is selected to update for each batch of training data. The data is first passed through the layers before the selected layer, then the selected layer and corresponding auxiliary classifier. Only parameters in the selected layer and its auxiliary head are updated. RLL maintains the main advantage of GLL in terms of plausibility, in that only one hidden layer is updated, requiring minimal error propagation, and, in addition, it does not require the sequential training of the network, which assumes strict timing of the layer updates.

\begin{algorithm}[t]
\caption{Difference Target Propagation with pooling layers (single-step)}
\begin{multicols}{2}
\begin{algorithmic}[1]\label{algo:dtp}
    \small
    \STATE Forward functions $f_l=\textsc{conv} + \textsc{pooling}$, $l=1,\dots,L$
    \STATE Backward functions $g_l = \textsc{strided deconv}$, $l=1,\dots,L-1$
    \STATE Input $\bfx_0$
    \STATE
    \FOR{$l=1$ to $L$} 
        \STATE {$\bfx_{l} = f_l(\bfx_{l-1})$} 
    \ENDFOR
    \STATE
    \FOR{$l=L$ to $2$}
        \STATE $\mathcal{L}_l^{inv} = \Vert g_l(f_l(\bfx_{l-1})) - \bfx_{l-1} \Vert_2^2$
        \STATE Update parameters in $g_l$ by minimizing $\mathcal{L}_l^{inv}$ using SGD
    \ENDFOR
    \STATE
    \STATE Set the first target: $\hat{\bfx}_{L} = \bfx_{L} - \eta_L\frac{\partial\mathcal{L}_L}{\partial \bfx_{L}}$
    
    \FOR{{$l=L$ to $2$}}
        \STATE $\hat{\bfx}_{l-1} = \bfx_{l-1} - g_l(\bfx_l) + g_l(\hat{\bfx}_l) $
    \ENDFOR
    \STATE
    \FOR{$l=1$ to $L$}
        \STATE $\mathcal{L}_l = \Vert f_l(\bfx_{l-1}) - \hat{\bfx}_l\Vert^2_2$ if $l<L$
        \STATE $\mathcal{L}_l = \mathcal{L}$ if $l=L$
        \STATE Update parameters in $f_l$ by minimizing $\mathcal{L}_l$ using SGD
    \ENDFOR

\end{algorithmic}
\end{multicols}
\end{algorithm}

\vspace{4mm}\noindent {\bf Incorporating Pooling Layers into Difference Target Propagation.} Target propagation (TP) circumvents the symmetric weight problem using layer-local losses. The main idea behind TP is to set a \textit{target} for each layer in the network, such that by reducing the distance between the feedforward activity and the feedback target in each layer, the global loss would be reduced as well. The targets are propagated top-down through a set of backward nonlinear functions trained using layer local autoencoders. Difference Target Propagation (DTP) is a variant of TP that introduces an error term into the backward propagation of targets, which accounts for the `imperfection' of the backward functions. A full description and proof of how and why DTP works can be found in \citet{lee2015difference} and \citet{MeulemansTP2020}.

In simple multi-layer perceptrons (MLPs), the forward functions $f_l$ and backward functions $g_l$ in DTP are simply linear layers with nonlinear activation functions, whereas in CNNs they have more complex structures. Convolutional layers are interleaved with pooling layers to perform down-sampling of the data, and thus $f_l$ is modelled as the combination of a convolutional layer and the subsequent pooling layers. \citet{bartunov2018assessing} claimed that pooling layers, which contain a deterministic step (either averaging or taking maximum), are incompatible with DTP, as the inverse of this many-to-one deterministic step can not be modelled using a simple nonlinear function $g_l$. Instead, they used strided convolutional layers to model $f_l$ and strided deconvolutional layers to model $g_l$ to perform down-sampling and the inverse up-sampling. However, we observe that we can retain pooling layers in the forward pass, i.e. $f_l = \textsc{conv} + \textsc{pooling}$, but use strided deconvolutional layers in the backward pass to approximate the inverse of the forward function, i.e. $g_l = \textsc{strided deconv}$. Essentially, this arrangement will enforce the strided deconvolution to learn the unpooling operation. A comparison of  the results in the Experiments section below demonstrates that incorporating max-pooling significantly improves the linear evaluation performances. The full algorithm for DTP with pooling layers is shown in Algorithm~\ref{algo:dtp}.

\section{Experiments}

\subsection{Linear evaluation on CIFAR10}\label{lin-eval-cifar}

\begin{table} [t]
\small
\centering 
\begin{tabular}{c c c c c} 
\toprule 
\textbf{Loss} & \textbf{Learning} & \textbf{Update} & \textbf{Acc.} & \textbf{+Grad. block and 5 negatives}\\
\midrule 
Contr. Hinge    & E2E & BP & 71.44\% & 70.35\% \\
                &     & DTP & 71.29\% & 67.74\%\\
                &     & RF & 61.70\% & 63.61\% \\
                & GLL & BP/URF & 72.76\% & 71.14\% \\
                &     & RF & 67.83\% & 66.52\% \\
                & RLL & BP/URF & 71.35\% & 71.17\% \\
                &     & RF & 65.94\% & 65.49\% \\
\hline
\textsc{SimCLR} & E2E & BP & 72.44\% & N/A\\
CLAPP & E2E & BP & 69.05\% & N/A \\
CLAPP & GLL & N/A & 68.93\% & N/A \\
\hline
Rnd. encoder & \multicolumn{4}{c}{61.23\%}\\ 
\bottomrule 
\end{tabular}
\caption{Test accuracy of a linear classifier trained on CIFAR10 embeddings from the SSL and CIFAR100-trained base encoder. We compare our contrastive hinge loss (Contr. Hinge) with $\textsc{SimCLR}$, and the encoder with randomly generated weights. We also compare the results from three different learning methods, End-to-End (E2E), Greedy Layer-wise Learning (GLL) and Randomized Layer-wise Learning (RLL), and four updating methods, back-propagation (BP), Updated Random Feedback (URF), Random Feedback (RF) and Difference Target Propagation (DTP). We further compared  the models with and without gradient block and a smaller number of negatives, as well as the CLAPP loss \citep{illing2021} with our deformations.} 
\label{tab:all-results} 
\end{table}

\noindent We first test the hypothesis that networks trained by our biologically learning rules produce embeddings with representational powers as good as those of networks trained by BP. The network architecture we use is as follows:

\begin{verbatim}
    Base Encoder:
    Conv 32 3x3 1; Hardtanh;
    Conv 32 3x3 1; Maxpool 2;
    Conv 64 3x3 1; Hardtanh;
    Conv 64 3x3 1; Maxpool 2;
    Conv 512 3x3 1; Maxpool 2;
    
    Projection Head:
    Flatten;
    Linear 64;
\end{verbatim}

\noindent where \verb+Conv 32 3x3 1+ stands for a convolutional layer with filter size 3, channel number 32 and stride 1, and \verb+Maxpool 2+ stands for a maxpooling layer with filter size 2 and stride size 2, which reduces the image size by a factor of 2. Before the final linear projector we simply flatten the feature space, applying no average pooling. We use CIFAR100 to train the base encoder. During training, each batch of the data will be randomly deformed twice to create the positive pairs (see Appendix \ref{image_exmaples} for examples of CIFAR100 deformations). All deformed images are passed into the base encoder and the encoder produces 64-dimensional embeddings, which are used to calculate the self-supervised loss. The parameters of the encoder are updated to minimize this loss. 

We use the standard linear evaluation scheme to evaluate the embeddings. After the base encoders are trained, we fix their parameters and remove
the projection head. We use the fixed layers to produce embeddings (the output of the last convolutional layer) as inputs to a linear classifier, trained on 45000 CIFAR10 examples and tested on 10000 CIFAR10 examples. The training of the linear classifier is supervised, and we use the classification accuracy as a measure of the representational power of the embeddings. We choose to use different datasets for training the encoders and classifiers because this is a more realistic learning scenario for biological systems: rather than learning from tasks on an ad hoc basis, the brain learns from more general tasks and data, and applies the learned synaptic weights to other tasks (e.g. classification). Our experiments with CIFAR10-trained base encoders on the same classification task yield identical performance.

\begin{figure}[t]
    \centering
    \noindent\makebox[\textwidth]{\includegraphics[width=0.9\paperwidth]{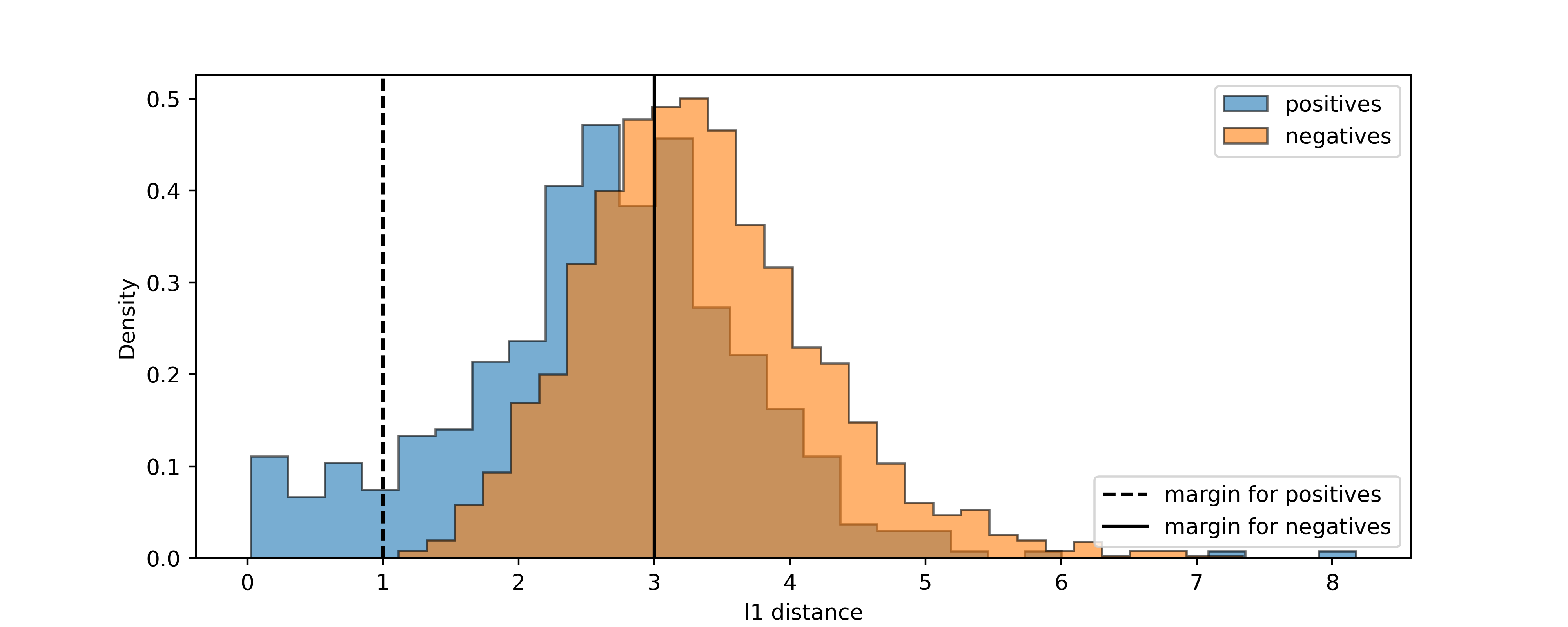}}
    
    \caption{The histograms of initial distances between positive pairs and negative pairs in an arbitrary batch of CIFAR100. Our loss selects positives corresponding to the right hand side of the dashed line, and negatives corresponding to the left hand side of the solid line.}
    \label{fig:init-dists}%
\end{figure}

\begin{figure}[t]
    \centering
    \noindent\makebox[\textwidth]{\includegraphics[width=0.9\paperwidth]{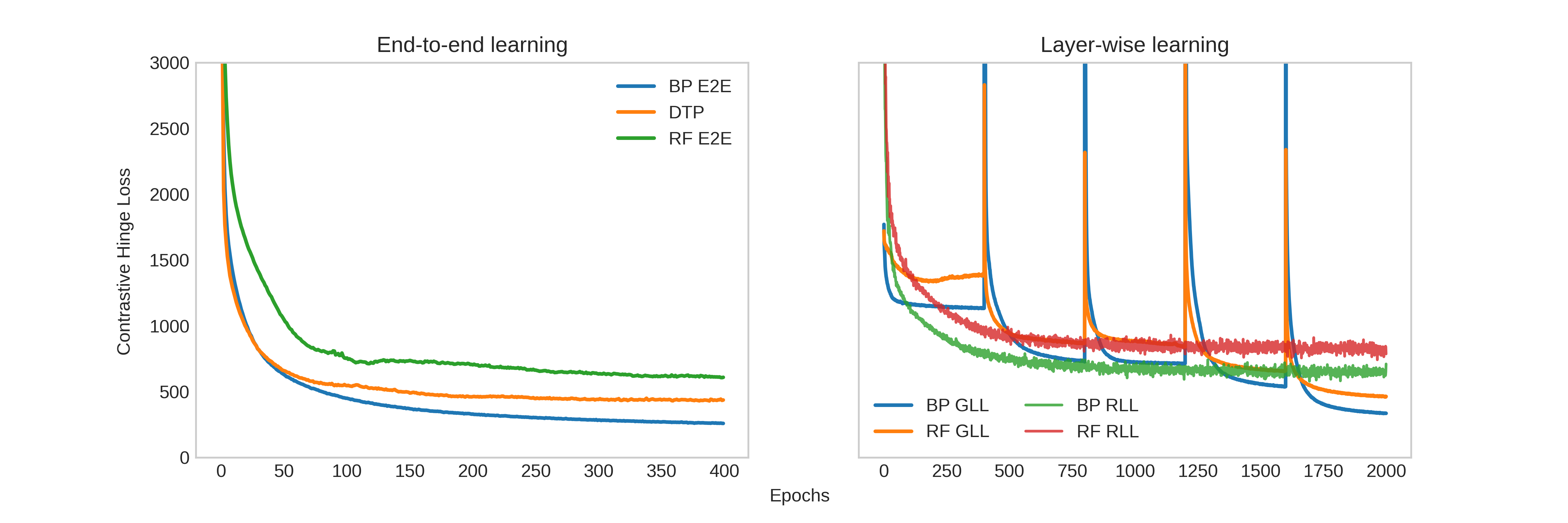}}
    
    \caption{Evolution of the contrastive hinge loss during training (excessively large values are cut off to maintain visibility). On the left, we plot the loss when the base encoder is trained end-to-end (E2E) with backpropagation (BP), random feedback (RF) and difference target propagation (DTP). One the right, we plot the loss when the base encoder is trained with greedy laryer-wise learning (GLL)/randomized layer-wise learning (RLL) and the weights are updated with BP/RF. For GLL, each layer is trained 400 epochs, and every 400 epochs in the plot corresponds to the training loss of a layer sequentially. For RLL, 5 layers are trained in a randomized way in a total of 2000 epochs, and the training loss shown is an average among all layers.}
    \label{fig:learning-curves}%
\end{figure}

For all experiments, we train the base encoder networks for 400 epochs using the Adam optimizer \citep{kingma2015}, with a learning rate 0.0001. Particularly, with DTP we use a fixed learning rate $0.001$ for the layer-wise autoencoders. For the contrastive losses we train the base encoders with a batch size of 500. The linear classifiers are trained using the Adam optimizer with a learning rate 0.001 for 400 epochs. For hyperparameters in the contrastive losses, we use $\tau=0.1$ for the \textsc{SimCLR} loss and $m_1=1,m_2=3$ for the contrastive hinge loss, based on the initial distance distributions between positives and negatives in an arbitrary batch (Figure~\ref{fig:init-dists}). Table~\ref{tab:all-results} shows our experimental results.  It can be seen that the proposed biologically more plausible contrastive hinge loss achieves similar performance to the \textsc{SimCLR} loss baseline. Overall, the biologically plausible learning rules achieve comparable linear evaluation performance to end-to-end BP. Layer-wise learning performs similarly to end-to-end learning with BP. There is some loss of accuracy with RF layer-wise learning but it still performs much better than end-to-end RF. In fact, the performance of end-to-end RF is close to that of a random encoder, which further demonstrates the failure of RF in deep networks. 
Notably, while earlier works with supervised learning \citep{bartunov2018assessing, lee2015difference} have shown a relatively large difference between BP and DTP in CIFAR classifications, in SSL we observe close performance between these two learning rules. Figure~\ref{fig:learning-curves} shows how the values of the contrastive hinge loss evolve using different training methods.  The results with $\textsc{SimCLR}$ combined with biologically plausible learning rules can be found in Appendix \ref{sec:appendix-cifar}.

In Table~\ref{tab:all-results} we also compared the linear evaluation performance of models with and without our proposed biological constraints, namely the gradient block and a smaller number of negatives. Models with the constraints achieve identical classification results to those without, although DTP experiences a slight performance drop compared to other biologically plausible training methods. Moreover by plugging the CLAPP loss proposed in  \citet{illing2021} into our deformation-based model and training it on CIFAR100, we obtain 69.05\% with end-to-end BP and 68.93\% with greedy layer-wise BP on CIFAR10.

\vspace{4mm}\noindent {\bf Pooling vs Strided Convolution in DTP.} We compare the linear evaluation performance of the encoder networks using two different down-sampling techniques, namely the max pooling layers and the strided convolutional layers, when the networks are trained by DTP. When using max pooling layers, the linear evaluation performance is 71.29\% for DTP (Table~\ref{tab:all-results}). However, this number quickly drops to 39.42\% with strided convolutional layers. Even for straightforward supervised classification training on CIFAR10, with the same architecture, there is a significant advantage using max pooling compared to strided convolution in conjunction with DTP, 71\% vs. 58\% . This result demonstrates that pooling layers can be compatible with DTP, and can also be better down-sampling techniques in certain architectures.

\begin{figure}[t]
    \centering
    \noindent\makebox[\textwidth]{\includegraphics[width=0.5\paperwidth]{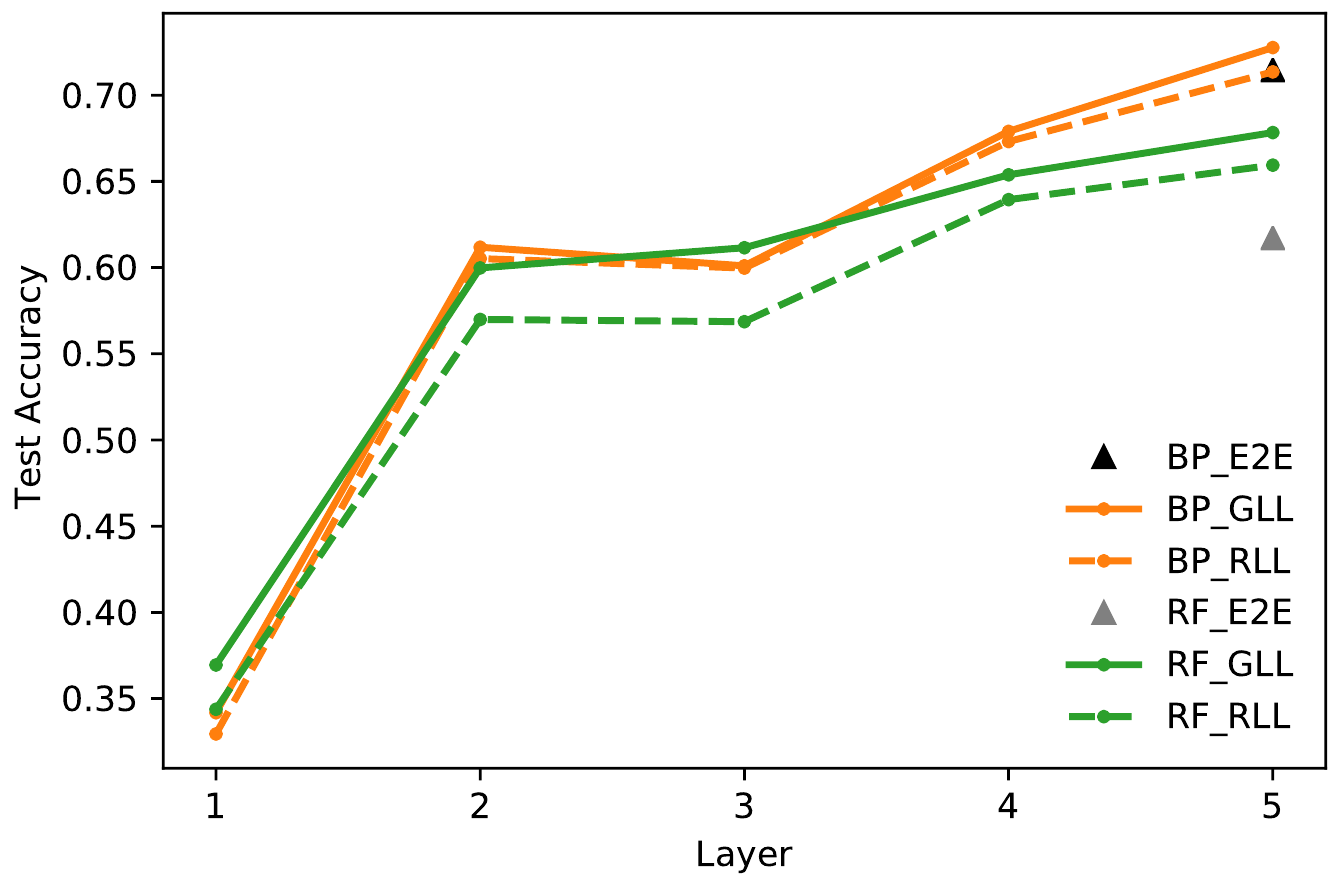}}
    
    \caption{CIFAR10 test accuracy of a linear classifier at every layer of an encoder network trained with contrastive hinge loss and End-to-End (E2E) learning, Greedy Layer-wise Learning (GLL) and Randomized Layer-wise Learning (RLL) and updating with Back-propagation (BP) and Random Feedback (RF)}%
    \label{fig:acc_layerwise}%
\end{figure}

\vspace{4mm}\noindent {\bf End-to-End and Layer-wise Learning.}
Figure~\ref{fig:acc_layerwise} compares the linear evaluation performance of end-to-end and layer-wise learning with BP/URF and RF, respectively. We can see that the layer-wise learning achieves similar, if not better, accuracy compared to the traditional end-to-end learning with BP updates. It is known that RF updates do not work well with deep networks when trained with end-to-end learning. Layer-wise learning with RF works around the problem by only updating one layer at a time and improves the performance by a large margin compared to end-to-end training with RF.

We examine the loss of layer-wise learning in Figure~\ref{fig:learning-curves} to further understand the behavior and convergence of the two layer-wise learning method. The networks are trained sequentially with GLL. We calculate the average training loss after each epoch, while a layer is trained. Every 400 epochs in the plot corresponds to the training loss of a layer sequentially. The training loss has a significant jump when a new layer is added to the top of the network, and then rapidly decreases to values lower than the previous layer and converges. For RLL, we train a total of 2000 epochs. Different layers are selected for each batch of data, and the training loss is an average among all batches, and hence all layers, in an epoch. The average training loss slightly oscillates and converges. Note that the values of the RLL loss cannot be directly compared to the loss of GLL since the training loss for RLL is an average across all layers.

The results show that layer-wise learning could be a more biologically plausible alternative to end-to-end learning. In addition, the effectiveness of RLL further indicates that it is not necessary to train the layers sequentially. Randomly selecting a layer to update and training all layers simultaneously gives competitive results compared to sequential layer-wise training.

\subsection{Linear evaluation on STL10}

For comparison with previous works \citep{lowe2019, illing2021}, we also perform our experiments with the STL10 dataset \citep{stl10}. We use the VGG6 network \citep{Simonyan15} , a VGG-like network that was used in \citet{illing2021}, but with a projection head of size 64 following a 2x2 average pooling layer. We first apply a 64x64 random resized crop on the original 96x96 images, and apply the other deformations in Figure~\ref{fig:deform2} to the images to create the positives and negatives. We take half of the unlabeled part of STL10, which contains 50000 data points, to train the base encoder for 100 epochs. When computing the contrastive hinge loss, we also use only 5 negative samples that are the closest to each positive pair within a batch. We have also performed preliminary experiments on models with 1 negative example, which yield similar performance to those with 5 negatives.  During linear evaluation, the classifier is trained on the 5000 training data points of labeled STL10 for 200 epochs, and tested on the 8000 testing data points of labeled STL10. We use a learning rate 0.0001 for training the base encoder, and 0.005 for training the classifier. We also experiment with the patch-based method and CLAPP loss proposed in \citet{illing2021}, where the positive pairs are created using vertical, adjacent 16x16 grayscale patches from the 64x64 crops. We experimented with the CLAPP loss with both end-to-end (E2E) learning and greedy layer-wise learning (GLL). The results are given in Table~\ref{tab:stl10}.

\begin{table} [t]
\small
\centering 
\begin{tabular}{c c c c} 
\toprule 
\textbf{Loss} & \textbf{Learning} & \textbf{Update} & \textbf{Acc.}\\
\midrule 
Contr. Hinge    & E2E & BP & 70.13\% \\
                &     & RF & 55.01\% \\
                & GLL & BP/URF & 68.26\% / \underline{68.80\%}\\
                &     & RF & 60.00\% / \underline{60.50\%}\\
                & RLL & BP/URF & 64.61\% \\
                &     & RF & 55.14\% \\
\hline
\textsc{CLAPP} & E2E & BP & 71.88\% \\
\textsc{CLAPP} & GLL & N/A & 68.74\% \\
\hline
Rnd. encoder & \multicolumn{3}{c}{46.78\%}\\ 
\bottomrule 
\end{tabular}
\caption{Linear evaluation results with labeled STL10, using a base encoder trained on unlabeled STL10. The contrastive hinge loss (Contr. Hinge) with 5 negatives is compared with patch-based CLAPP in both end-to-end (E2E) and greedy layer-wise learning (GLL). We also compare the results from randomized layer-wise learning (RLL), and four updating methods, backpropagation (BP), updated random feedback (URF), random feedback (RF) when using contrastive hinge loss. Underlined results were obtained with 1 negative.} 
\label{tab:stl10} 
\end{table}

\begin{figure}[t]
    \centering
    \noindent\makebox[\textwidth]{\includegraphics[width=0.7\paperwidth]{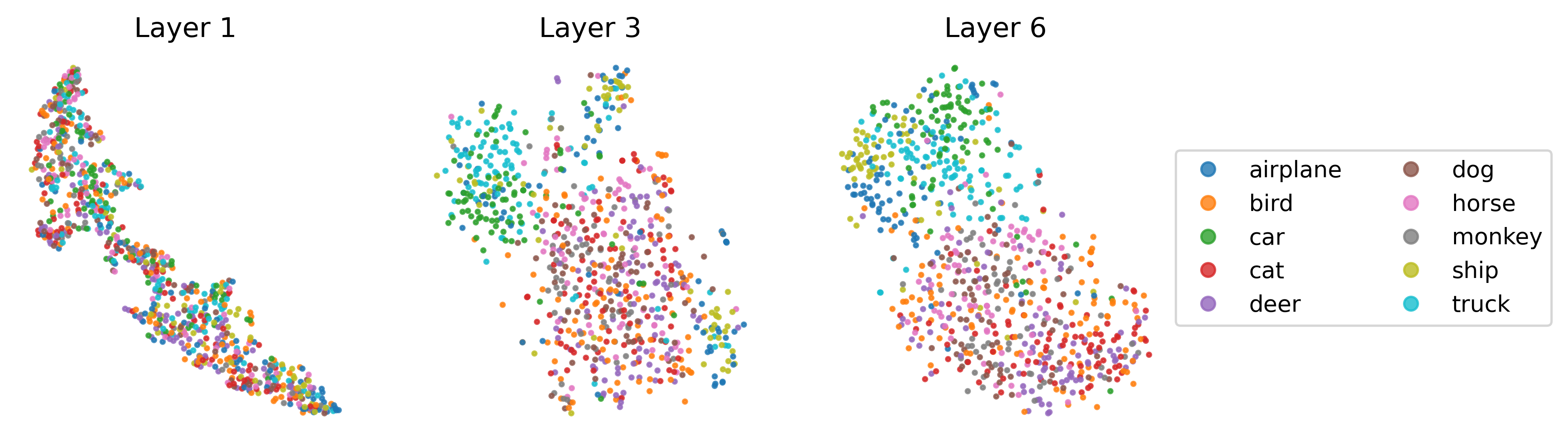}}
    
    \caption{t-SNE visualization of the embeddings of STL10 labeled test data, from a VGG6 network trained by greedy layer-wise learning (GLL) using BP only for the 2-layers involved in GLL. We apply global average pooling to embeddings to get vector representations and then apply t-SNE. Each dot represents an image.}
    \label{fig:tsne}%
\end{figure}

To understand whether the network discovers more semantic features of the input images as the network goes deeper, we visualize in Figure~\ref{fig:tsne} the embeddings of the test set in labeled STL10 using the dimensionality reduction technique t-SNE \citep{van2008visualizing}, in each layer of a network trained by BP and greedy layer-wise learning. As the network deepens, the separation between embeddings representing `animals' and `vehicles' becomes increasingly obvious, and the clustering effect of each specific category within the two large categories also strengthens, especially on the `vehicles' side. Notice that at this stage, the downstream classifier has not been trained and the base encoder has not seen the labeled data, suggesting that the unsupervised learning alone has already enabled the network to develop certain levels of semantic understanding of the data.

\subsection{Handling variability}

\begin{figure}[t]
    \centering
    \noindent\makebox[\textwidth]{\includegraphics[width=0.5\paperwidth]{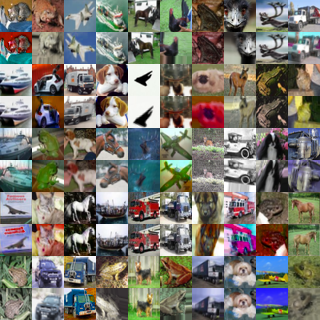}}
    
    \caption{Examples of the deformed CIFAR10 images (even rows) used to test the classifier trained on normal, undeformed CIFAR10 (odd rows).}
    \label{fig:deformed_cifar}%
\end{figure}

In order to explore the robustness of our SSL method to perturbations unseen during training, we design the following experiments. First, we train a base encoder using CIFAR100 and the network architecture mentioned in Section~\ref{lin-eval-cifar} above, with our contrastive hinge loss. We then use this base encoder to produce embeddings of CIFAR10, and train a classifier on these embeddings. We then test the classifier on a set of deformed CIFAR10 images shown in Figure~\ref{fig:deformed_cifar}. We also train a supervised network with the same architecture on `normal' CIFAR10 and test it on deformed images. For both self-supervised and supervised networks we experiment with 5000 and 50000 training data for the classifier. The results in Table~\ref{tab:cifar_deformed} suggest that self-supervised networks with the contrastive hinge loss are indeed more robust to deformations or perturbations than supervised networks, even with a limited amount of data. In the supervised setting it is possible to use data augmentation to improve the robustness, but what we see here is that by using the robust embedding it is possible to quickly learn robust classifiers without the extensive data augmentation process for each subsequent classification task. 

\begin{table} [t]
\small
\centering 
\begin{tabular}{c | c | c } 
\toprule 
\textbf{Size of training data} & \textbf{SSL with Contr. Hinge} & \textbf{Supervised} \\
\midrule
{\bf 50000}   & 59.34\% & 53.83\% \\
{\bf 5000}    & 52.84\% & 43.27\% \\
\bottomrule 
\end{tabular}
\caption{Test accuracy on deformed CIFAR10 images, using a single-layer classifier trained with SSL embeddings or a network with supervised learning directly. Training data refers to `normal' CIFAR10 images. } 
\label{tab:cifar_deformed} 
\end{table}

For further explorations of the perturbation robustness property of SSL combined with biologically plausible learning, we use two other sets of data, namely the 28x28 gray-scale digits (MNIST) \citep{lecun-mnisthandwrittendigit-2010} and letters (EMNIST) \citep{cohen2017emnist}. We first train the base encoders on EMNIST, and then we use the trained encoders to produce embeddings of MNIST and train a linear classifier on these embeddings. We then test the classifiers on a set of affine-transformed MNIST. Examples of the original and transformed EMNIST and MNIST can be found in Appendix \ref{image_exmaples}. This experiment design aims to verify the robustness of the SSL framework, by comparing its performance with a fully supervised CNN trained by MNIST and tested on transformed MNIST. For this set of experiments, we use a shallower CNN with the following architecture:

\begin{verbatim}
    Base Encoder:
    Conv 32 3x3 1; Maxpool 2; Tanh;
    Conv 64 3x3 1; Maxpool 2; Tanh;
    Conv 128 3x3 1; Maxpool 2; Tanh;
    
    Projection Head:
    Flatten;
    Linear 64;
\end{verbatim}

We note that for the relatively easy task of classifying MNIST digits, all the unsupervised training methods perform above 98\%, even using end to end feedback alignment. After all a random network with this structure, where we only train the linear classification layer, also yields above 98\% accuracy. Thus the interesting question is how well does the unsupervised network generalize to the transformed digits.

We train the base encoder networks for 400 epochs using the Adam optimizer, with a learning rate chosen between $[0.0001,0.001]$. In DTP, we again use a fixed learning rate 0.001 for the layer-wise auto-encoders.  We train the base encoders with a batch size 1000, and the linear classifier is trained for 200 epochs using Adam optimizer with a learning rate of 0.001. We use $m_1=1, m_2=1.5$ for the contrastive hinge loss. The results of these experiments are shown in Table~\ref{tab:mnists}.

\begin{table} [t]
\small
\centering 
\begin{tabular}{c c c c} 
\toprule 
\textbf{Loss} & \textbf{Learning} & \textbf{Update} & \textbf{Acc. on transformed digits}\\
\midrule
Contr. Hinge    & E2E & BP & 80.20\% \\
                &     & DTP & 81.37\% \\
                &     & RF & 75.89\% \\
                & GLL & BP & 81.31\% \\
                &     & RF & 80.86\% \\
                & RLL & BP & 80.31\% \\
                &     & RF & 78.52\% \\
\midrule
\textsc{SimCLR} & E2E & BP & 81.96\% \\
\hline
Supervised & \multicolumn{3}{c}{75.6\%}\\
Rnd. encoder & \multicolumn{3}{c}{62.66\%}\\ 
\bottomrule 
\end{tabular}
\caption{Test accuracy on transformed MNIST using the pre-trained networks with SSL on EMNIST letters. Linear classifiers/supervised networks trained on original MNIST.} 
\label{tab:mnists} 
\end{table}

We found that all classification results on the the transformed MNIST by SSL are better than those by training a supervised network with MNIST digits and testing it on transformed MNIST digits directly, showing the robustness of our proposed biologically more plausible SSL method. Since traditionally robustness of classification ANNs is ensured by applying data augmentations to the training data, our results suggest that there may not be a need to augment the data when training the classifiers, making learning more efficient.

\section{Discussion}

In this work we have shown that it is possible to construct a biologically plausible deep learning framework, using local Hebbian updates and in some cases avoiding the locking problem, using self-supervised learning (SSL) combined with layer-wise learning or difference target propagation (DTP). We have also introduced a very simple  contrastive loss whose gradient involves only local updates based on pre and post synaptic activities. An embedding is learned through the self-supervised contrastive loss using a large collection of unlabelled images. The evaluation of the embedding is done by training a linear classifier, taking as input the trained embedding, but using a separate collection of labeled images from different categories. 
Layer-wise learning requires physical connections between each layer and the error computing layer. This is consistent with the fact that direct connections do exist between various retinotopic layers and higher cortical areas \citep{VanEssen}. And the fact that randomized layer-wise training is effective means that there is no need to sequence the learning of the different layers.

Using these alternative learning methods, we have produced comparable embeddings to the backprop-trained ones for linear evaluation. The small reduction in downstream accuracy 
would easily be remedied by the essentially unlimited amount of unlabeled data available to the visual system. We have shown that the perturbations inherent in the self-supervised learning yield embeddings that are more robust to perturbations than direct classifier training. Furthermore we have shown that there is no need for large numbers of negative examples for each positive pair in order to obtain the same results. A future direction of research would involve actually training on continuous videos of moving objects, with small buffers for the negative images.

The SSL methods that avoid computing a contrast with negative examples are of particular interest. They try to force the embedding of the positives to be `spread out' over the embedding space. The related losses that have been proposed in the literature involve rather complex computations, and in further research we would like to explore biologically plausible alternatives of them.

Considering the approach of using ANNs as models of cortical areas in general, although \citet{yamins2016using} have shown that in many way CNNs are similar to the ventral visual stream, including the end-to-end behavior and the population level activities, many aspects of CNNs are still far from biologically realistic. For example, units in CNNs are continuously valued, while real neurons are discrete, emitting binary spikes. Even if we use CNNs to simply model the firing rates of neurons, negative unit values are still a problem. Another source of implausibility is that CNNs involve only feedforward connections, while biological neurons in the cortex are also recurrently connected, with different connectivity patterns between different cell types and areas. Moreover, the weight sharing property of convolutional layers lacks neurobiological support, and using locally connected layers \citep{bartunov2018assessing,amit2019deep} may be a more biologically plausible approach. To further develop our proposed framework, these aspects are definitely to be considered.

\bigskip

{\bf Acknowledgements:} We thank the anonymous referees for important and helpful suggestions. This work was supported in part by NIMH/CRCNS award no. R01 MH11555. 
\newpage

\clearpage

\typeout{}
\bibliographystyle{agsm}
\bibliography{refs}

@article {VanEssen,
	author = {Van Essen, David C. and Donahue, Chad J. and Coalson, Timothy S. and Kennedy, Henry and Hayashi, Takuya and Glasser, Matthew F.},
	title = {Cerebral cortical folding, parcellation, and connectivity in humans, nonhuman primates, and mice},
	volume = {116},
	number = {52},
	pages = {26173--26180},
	year = {2019},
	doi = {10.1073/pnas.1902299116},
	publisher = {National Academy of Sciences},
	issn = {0027-8424},
	URL = {https://www.pnas.org/content/116/52/26173},
	eprint = {https://www.pnas.org/content/116/52/26173.full.pdf},
	journal = {Proceedings of the National Academy of Sciences}
}

@inproceedings{lee2015difference,
  title={Difference target propagation},
  author={Lee, Dong-Hyun and Zhang, Saizheng and Fischer, Asja and Bengio, Yoshua},
  booktitle={Joint european conference on machine learning and knowledge discovery in databases},
  pages={498--515},
  year={2015},
  organization={Springer}
}

@inproceedings{akrout2019deep,
 author = {Akrout, Mohamed and Wilson, Collin and Humphreys, Peter and Lillicrap, Timothy and Tweed, Douglas B},
 booktitle = {Advances in Neural Information Processing Systems},
 editor = {H. Wallach and H. Larochelle and A. Beygelzimer and F. d\textquotesingle Alch\'{e}-Buc and E. Fox and R. Garnett},
 pages = {},
 publisher = {Curran Associates, Inc.},
 title = {Deep Learning without Weight Transport},
 url = {https://proceedings.neurips.cc/paper/2019/file/f387624df552cea2f369918c5e1e12bc-Paper.pdf},
 volume = {32},
 year = {2019}
}

@article{Lilli,
  title={Random synaptic feedback weights support error backpropagation for deep learning},
  author={Lillicrap, Timothy P and Cownden, Daniel and Tweed, Douglas B and Akerman, Colin J},
  journal={Nature communications},
  volume={7},
  number={1},
  pages={1--10},
  year={2016},
  publisher={Nature Publishing Group}
}

@inproceedings{chen2020simple,
  title={A simple framework for contrastive learning of visual representations},
  author={Chen, Ting and Kornblith, Simon and Norouzi, Mohammad and Hinton, Geoffrey},
  booktitle={International conference on machine learning},
  pages={1597--1607},
  year={2020},
  organization={PMLR}
}

@inproceedings{he2020momentum,
  title={Momentum contrast for unsupervised visual representation learning},
  author={He, Kaiming and Fan, Haoqi and Wu, Yuxin and Xie, Saining and Girshick, Ross},
  booktitle={Proceedings of the IEEE/CVF Conference on Computer Vision and Pattern Recognition},
  pages={9729--9738},
  year={2020}
}

@incollection{zipser1993neurobiological,
  title={The neurobiological significance of the new learning models},
  author={Zipser, David and Rumelhart, David E},
  booktitle={Computational neuroscience},
  pages={192--200},
  year={1993}
}

@article{rao1999predictive,
  title={Predictive coding in the visual cortex: a functional interpretation of some extra-classical receptive-field effects},
  author={Rao, Rajesh PN and Ballard, Dana H},
  journal={Nature neuroscience},
  volume={2},
  number={1},
  pages={79--87},
  year={1999},
  publisher={Nature Publishing Group}
}

@article{oord2018representation,
  title={Representation learning with contrastive predictive coding},
  author={Oord, Aaron van den and Li, Yazhe and Vinyals, Oriol},
  journal={arXiv preprint arXiv:1807.03748},
  year={2018}
}

@inproceedings{grill2020bootstrap,
 author = {Grill, Jean-Bastien and Strub, Florian and Altch\'{e}, Florent and Tallec, Corentin and Richemond, Pierre and Buchatskaya, Elena and Doersch, Carl and Avila Pires, Bernardo and Guo, Zhaohan and Gheshlaghi Azar, Mohammad and Piot, Bilal and kavukcuoglu, koray and Munos, Remi and Valko, Michal},
 booktitle = {Advances in Neural Information Processing Systems},
 editor = {H. Larochelle and M. Ranzato and R. Hadsell and M. F. Balcan and H. Lin},
 pages = {21271--21284},
 publisher = {Curran Associates, Inc.},
 title = {Bootstrap Your Own Latent - A New Approach to Self-Supervised Learning},
 url = {https://proceedings.neurips.cc/paper/2020/file/f3ada80d5c4ee70142b17b8192b2958e-Paper.pdf},
 volume = {33},
 year = {2020}
}

@article{zbontar2021barlow,
  title={Barlow Twins: Self-Supervised Learning via Redundancy Reduction},
  author={Zbontar, Jure and Jing, Li and Misra, Ishan and LeCun, Yann and Deny, St{\'e}phane},
  journal={arXiv preprint arXiv:2103.03230},
  year={2021}
}

@inproceedings{chen2020exploring,
  title={Exploring simple siamese representation learning},
  author={Chen, Xinlei and He, Kaiming},
  booktitle={Proceedings of the IEEE/CVF Conference on Computer Vision and Pattern Recognition},
  pages={15750--15758},
  year={2021}
}

@article{zhuang2021unsupervised,
  title={Unsupervised neural network models of the ventral visual stream},
  author={Zhuang, Chengxu and Yan, Siming and Nayebi, Aran and Schrimpf, Martin and Frank, Michael C and DiCarlo, James J and Yamins, Daniel LK},
  journal={Proceedings of the National Academy of Sciences},
  volume={118},
  number={3},
  year={2021},
  publisher={National Acad Sciences}
}

@article{schrimpf2018brain,
  title={Brain-score: Which artificial neural network for object recognition is most brain-like?},
  author={Schrimpf, Martin and Kubilius, Jonas and Hong, Ha and Majaj, Najib J and Rajalingham, Rishi and Issa, Elias B and Kar, Kohitij and Bashivan, Pouya and Prescott-Roy, Jonathan and Schmidt, Kailyn and others},
  journal={BioRxiv},
  pages={407007},
  year={2018},
  publisher={Cold Spring Harbor Laboratory}
}

@article{carandini2012normalization,
  title={Normalization as a canonical neural computation},
  author={Carandini, Matteo and Heeger, David J},
  journal={Nature Reviews Neuroscience},
  volume={13},
  number={1},
  pages={51--62},
  year={2012},
  publisher={Nature Publishing Group}
}

@article{krizhevsky2009learning,
  title={The cifar-10 dataset},
  author={Krizhevsky, Alex and Nair, Vinod and Hinton, Geoffrey},
  journal={online: http://www. cs. toronto. edu/kriz/cifar. html},
  volume={55},
  number={5},
  year={2014}
}

@inproceedings{cohen2017emnist,
  title={EMNIST: Extending MNIST to handwritten letters},
  author={Cohen, Gregory and Afshar, Saeed and Tapson, Jonathan and Van Schaik, Andre},
  booktitle={2017 International Joint Conference on Neural Networks (IJCNN)},
  pages={2921--2926},
  year={2017},
  organization={IEEE}
}

@article{barlow1961possible,
  title={Possible principles underlying the transformation of sensory messages},
  author={Barlow, Horace B and others},
  journal={Sensory communication},
  volume={1},
  number={01},
  year={1961}
}

@article{lecun-mnisthandwrittendigit-2010,
  title={MNIST handwritten digit database. 2010},
  author={LeCun, Yann and Cortes, Corinna and Burges, Christopher J},
  journal={URL http://yann. lecun. com/exdb/mnist},
  volume={7},
  number={23},
  pages={6},
  year={2010}
}

@inproceedings{bartunov2018assessing,
 author = {Bartunov, Sergey and Santoro, Adam and Richards, Blake and Marris, Luke and Hinton, Geoffrey E and Lillicrap, Timothy},
 booktitle = {Advances in Neural Information Processing Systems},
 editor = {S. Bengio and H. Wallach and H. Larochelle and K. Grauman and N. Cesa-Bianchi and R. Garnett},
 pages = {},
 publisher = {Curran Associates, Inc.},
 title = {Assessing the Scalability of Biologically-Motivated Deep Learning Algorithms and Architectures},
 url = {https://proceedings.neurips.cc/paper/2018/file/63c3ddcc7b23daa1e42dc41f9a44a873-Paper.pdf},
 volume = {31},
 year = {2018}
}

@article{yamins2016using,
  title={Using goal-driven deep learning models to understand sensory cortex},
  author={Yamins, Daniel LK and DiCarlo, James J},
  journal={Nature neuroscience},
  volume={19},
  number={3},
  pages={356--365},
  year={2016},
  publisher={Nature Publishing Group}
}

@article{mcintosh2016deep,
  title={Deep learning models of the retinal response to natural scenes},
  author={McIntosh, Lane T and Maheswaranathan, Niru and Nayebi, Aran and Ganguli, Surya and Baccus, Stephen A},
  journal={Advances in neural information processing systems},
  volume={29},
  pages={1369},
  year={2016},
  publisher={NIH Public Access}
}

@article{masse2019circuit,
  title={Circuit mechanisms for the maintenance and manipulation of information in working memory},
  author={Masse, Nicolas Y and Yang, Guangyu R and Song, H Francis and Wang, Xiao-Jing and Freedman, David J},
  journal={Nature neuroscience},
  volume={22},
  number={7},
  pages={1159--1167},
  year={2019},
  publisher={Nature Publishing Group}
}

@article{song2016training,
  title={Training excitatory-inhibitory recurrent neural networks for cognitive tasks: a simple and flexible framework},
  author={Song, H Francis and Yang, Guangyu R and Wang, Xiao-Jing},
  journal={PLoS computational biology},
  volume={12},
  number={2},
  pages={e1004792},
  year={2016},
  publisher={Public Library of Science San Francisco, CA USA}
}

@article{rumelhart1986learning,
  title={Learning representations by back-propagating errors},
  author={Rumelhart, David E and Hinton, Geoffrey E and Williams, Ronald J},
  journal={nature},
  volume={323},
  number={6088},
  pages={533--536},
  year={1986},
  publisher={Nature Publishing Group}
}

@article{amit2019deep,
  title={Deep learning with asymmetric connections and Hebbian updates},
  author={Amit, Yali},
  journal={Frontiers in computational neuroscience},
  volume={13},
  pages={18},
  year={2019},
  publisher={Frontiers}
}

@article{lillicrap2020backpropagation,
  title={Backpropagation and the brain},
  author={Lillicrap, Timothy P and Santoro, Adam and Marris, Luke and Akerman, Colin J and Hinton, Geoffrey},
  journal={Nature Reviews Neuroscience},
  volume={21},
  number={6},
  pages={335--346},
  year={2020},
  publisher={Nature Publishing Group}
}

@article{Becker1992SelforganizingNN,
  title={Self-organizing neural network that discovers surfaces in random-dot stereograms},
  author={S. Becker and Geoffrey E. Hinton},
  journal={Nature},
  year={1992},
  volume={355},
  pages={161-163}
}

@article{whittington2017approximation,
  title={An approximation of the error backpropagation algorithm in a predictive coding network with local hebbian synaptic plasticity},
  author={Whittington, James CR and Bogacz, Rafal},
  journal={Neural computation},
  volume={29},
  number={5},
  pages={1229--1262},
  year={2017},
  publisher={MIT Press}
}

@inproceedings{liao2016important,
  title={How important is weight symmetry in backpropagation?},
  author={Liao, Qianli and Leibo, Joel and Poggio, Tomaso},
  booktitle={Proceedings of the AAAI Conference on Artificial Intelligence},
  volume={30},
  number={1},
  year={2016}
}

@inproceedings{bengio2007greedy,
  title={Greedy layer-wise training of deep networks},
  author={Bengio, Yoshua and Lamblin, Pascal and Popovici, Dan and Larochelle, Hugo},
  booktitle={Advances in neural information processing systems},
  pages={153--160},
  year={2007}
}

@article{hinton2006fast,
  title={A fast learning algorithm for deep belief nets},
  author={Hinton, Geoffrey E and Osindero, Simon and Teh, Yee-Whye},
  journal={Neural computation},
  volume={18},
  number={7},
  pages={1527--1554},
  year={2006},
  publisher={MIT Press One Rogers Street, Cambridge, MA 02142-1209, USA journals-info~…}
}

@InProceedings{huang2018learning,
  title = 	 {Learning Deep {R}es{N}et Blocks Sequentially using Boosting Theory},
  author =       {Huang, Furong and Ash, Jordan and Langford, John and Schapire, Robert},
  booktitle = 	 {Proceedings of the 35th International Conference on Machine Learning},
  pages = 	 {2058--2067},
  year = 	 {2018},
  editor = 	 {Dy, Jennifer and Krause, Andreas},
  volume = 	 {80},
  series = 	 {Proceedings of Machine Learning Research},
  month = 	 {10--15 Jul},
  publisher =    {PMLR},
  url = 	 {https://proceedings.mlr.press/v80/huang18b.html},
}

@inproceedings{belilovskyGreedyLayerwiseLearning2019,
  title={Greedy layerwise learning can scale to imagenet},
  author={Belilovsky, Eugene and Eickenberg, Michael and Oyallon, Edouard},
  booktitle={International conference on machine learning},
  pages={583--593},
  year={2019},
  organization={PMLR}
}

@InProceedings{jaderbergDecoupledNeuralInterfaces2017,
  title = 	 {Decoupled Neural Interfaces using Synthetic Gradients},
  author =       {Max Jaderberg and Wojciech Marian Czarnecki and Simon Osindero and Oriol Vinyals and Alex Graves and David Silver and Koray Kavukcuoglu},
  booktitle = 	 {Proceedings of the 34th International Conference on Machine Learning},
  pages = 	 {1627--1635},
  year = 	 {2017},
  editor = 	 {Precup, Doina and Teh, Yee Whye},
  volume = 	 {70},
  series = 	 {Proceedings of Machine Learning Research},
  month = 	 {06--11 Aug},
  publisher =    {PMLR},
  url = 	 {https://proceedings.mlr.press/v70/jaderberg17a.html},
}

@article{tsodyks1,
author = {Gianluigi Mongillo  and Omri Barak  and Misha Tsodyks },
title = {Synaptic Theory of Working Memory},
journal = {Science},
volume = {319},
number = {5869},
pages = {1543-1546},
year = {2008},
doi = {10.1126/science.1150769},

URL = {https://www.science.org/doi/abs/10.1126/science.1150769},
eprint = {https://www.science.org/doi/pdf/10.1126/science.1150769}
}

@article{tsodyks2,
title = {Working models of working memory},
journal = {Current Opinion in Neurobiology},
volume = {25},
pages = {20-24},
year = {2014},
note = {Theoretical and computational neuroscience},
issn = {0959-4388},
doi = {https://doi.org/10.1016/j.conb.2013.10.008},
url = {https://www.sciencedirect.com/science/article/pii/S0959438813002158},
author = {Omri Barak and Misha Tsodyks}
}

@InProceedings{noklandTrainingNeuralNetworks2019,
  title = 	 {Training Neural Networks with Local Error Signals},
  author =       {N{\o}kland, Arild and Eidnes, Lars Hiller},
  booktitle = 	 {Proceedings of the 36th International Conference on Machine Learning},
  pages = 	 {4839--4850},
  year = 	 {2019},
  editor = 	 {Chaudhuri, Kamalika and Salakhutdinov, Ruslan},
  volume = 	 {97},
  series = 	 {Proceedings of Machine Learning Research},
  month = 	 {09--15 Jun},
  publisher =    {PMLR},
  url = 	 {https://proceedings.mlr.press/v97/nokland19a.html},
}

@article{ororbia2020large,
  title={Large-Scale Gradient-Free Deep Learning with Recursive Local Representation Alignment},
  author={Ororbia, Alexander and Mali, Ankur and Kifer, Daniel and Giles, C Lee},
  journal={arXiv preprint arXiv:2002.03911},
  year={2020}
}

@inproceedings{illing2021,
  title={Local plasticity rules can learn deep representations using self-supervised contrastive predictions},
  author={Illing, Bernd and Ventura, Jean Robin and Bellec, Guillaume and Gerstner, Wulfram},
  booktitle={Thirty-Fifth Conference on Neural Information Processing Systems},
  year={2021}
}

@inproceedings{swav2020,
 author = {Caron, Mathilde and Misra, Ishan and Mairal, Julien and Goyal, Priya and Bojanowski, Piotr and Joulin, Armand},
 booktitle = {Advances in Neural Information Processing Systems},
 editor = {H. Larochelle and M. Ranzato and R. Hadsell and M. F. Balcan and H. Lin},
 pages = {9912--9924},
 publisher = {Curran Associates, Inc.},
 title = {Unsupervised Learning of Visual Features by Contrasting Cluster Assignments},
 url = {https://proceedings.neurips.cc/paper/2020/file/70feb62b69f16e0238f741fab228fec2-Paper.pdf},
 volume = {33},
 year = {2020}
}

@inproceedings{lowe2019,
 author = {L\"{o}we, Sindy and O\textquotesingle Connor, Peter and Veeling, Bastiaan},
 booktitle = {Advances in Neural Information Processing Systems},
 editor = {H. Wallach and H. Larochelle and A. Beygelzimer and F. d\textquotesingle Alch\'{e}-Buc and E. Fox and R. Garnett},
 pages = {},
 publisher = {Curran Associates, Inc.},
 title = {Putting An End to End-to-End: Gradient-Isolated Learning of Representations},
 url = {https://proceedings.neurips.cc/paper/2019/file/851300ee84c2b80ed40f51ed26d866fc-Paper.pdf},
 volume = {32},
 year = {2019}
}

@inproceedings{ganguli2021,
  author    = {Yuandong Tian and
               Xinlei Chen and
               Surya Ganguli},
  editor    = {Marina Meila and
               Tong Zhang},
  title     = {Understanding self-supervised learning dynamics without contrastive
               pairs},
  booktitle = {Proceedings of the 38th International Conference on Machine Learning,
               {ICML} 2021, 18-24 July 2021, Virtual Event},
  series    = {Proceedings of Machine Learning Research},
  volume    = {139},
  pages     = {10268--10278},
  publisher = {{PMLR}},
  year      = {2021},
  url       = {http://proceedings.mlr.press/v139/tian21a.html},
  bibsource = {dblp computer science bibliography, https://dblp.org}
}

@book{hebb1949organization,
  title={The {O}rganization of {B}ehavior.},
  author={Hebb, Donald},
  year={1949},
  publisher={Wiley},
  address={New York}
}

@inproceedings{henaff2020data,
  title={Data-efficient image recognition with contrastive predictive coding},
  author={Henaff, Olivier},
  booktitle={International Conference on Machine Learning},
  pages={4182--4192},
  year={2020},
  organization={PMLR}
}

@article{manchev2020target,
  title={Target Propagation in Recurrent Neural Networks.},
  author={Manchev, Nikolay and Spratling, Michael W},
  journal={J. Mach. Learn. Res.},
  volume={21},
  pages={7--1},
  year={2020}
}

@inproceedings{MeulemansTP2020,
 author = {Meulemans, Alexander and Carzaniga, Francesco and Suykens, Johan and Sacramento, Jo\~{a}o and Grewe, Benjamin F.},
 booktitle = {Advances in Neural Information Processing Systems},
 editor = {H. Larochelle and M. Ranzato and R. Hadsell and M. F. Balcan and H. Lin},
 pages = {20024--20036},
 publisher = {Curran Associates, Inc.},
 title = {A Theoretical Framework for Target Propagation},
 url = {https://proceedings.neurips.cc/paper/2020/file/e7a425c6ece20cbc9056f98699b53c6f-Paper.pdf},
 volume = {33},
 year = {2020}
}

@InProceedings{stl10,
  title = 	 {An Analysis of Single-Layer Networks in Unsupervised Feature Learning},
  author = 	 {Coates, Adam and Ng, Andrew and Lee, Honglak},
  booktitle = 	 {Proceedings of the Fourteenth International Conference on Artificial Intelligence and Statistics},
  pages = 	 {215--223},
  year = 	 {2011},
  editor = 	 {Gordon, Geoffrey and Dunson, David and Dudík, Miroslav},
  volume = 	 {15},
  series = 	 {Proceedings of Machine Learning Research},
  address = 	 {Fort Lauderdale, FL, USA},
  month = 	 {11--13 Apr},
  publisher =    {PMLR},
  url = 	 {https://proceedings.mlr.press/v15/coates11a.html},
}

@InProceedings{Simonyan15,
  author       = "Karen Simonyan and Andrew Zisserman",
  title        = "Very Deep Convolutional Networks for Large-Scale Image Recognition",
  booktitle    = "International Conference on Learning Representations",
  year         = "2015",
}

@article{van2008visualizing,
  title={Visualizing data using t-SNE.},
  author={Van der Maaten, Laurens and Hinton, Geoffrey},
  journal={Journal of machine learning research},
  volume={9},
  number={11},
  year={2008}
}

@inproceedings{kingma2015,
  author    = {Diederik P. Kingma and
               Jimmy Ba},
  editor    = {Yoshua Bengio and
               Yann LeCun},
  title     = {Adam: {A} Method for Stochastic Optimization},
  booktitle = {3rd International Conference on Learning Representations, {ICLR} 2015,
               San Diego, CA, USA, May 7-9, 2015, Conference Track Proceedings},
  year      = {2015},
  bibsource = {dblp computer science bibliography, https://dblp.org}
}

\appendix

\newpage%
\renewcommand{\thesection}{\Alph{section}}
\section{Computing the thresholds for the error signal at the top layer}\label{threshold}
For each unit $i$ in layer $L$
one can imagine using an inhibitory unit $u^I_i$
activated by $x^{B}_{L,i}$ and  another unit $x^{D,+}_i=x^{A}_{L,i}+u^I_i$. Similarly an inhibitory unit $v^I_i$ is activated by
$x^{A}_{L,i}$ and $x^{D,-} = v^I_i + x^{B}_{L,i}$.
Finally two units compute the thresholded sums:
$x^{S,+}= \sum_i x^{D,+}_i > m_1, x^{S,-}= \sum_i x^{D,-}_i > m_2$, and if any of these is above threshold
the quantity in $x^{D,+}_i$ is passed on as $\delta^B_{L,i}$.
A similar architecture can be used to compute $\delta^B_{t,L,i}$ for each negative $t$.

\section{Examples of deformations}\label{image_exmaples}

\begin{figure}[t]
    \centering
    \noindent\makebox[\textwidth]{\includegraphics[width=0.6\paperwidth]{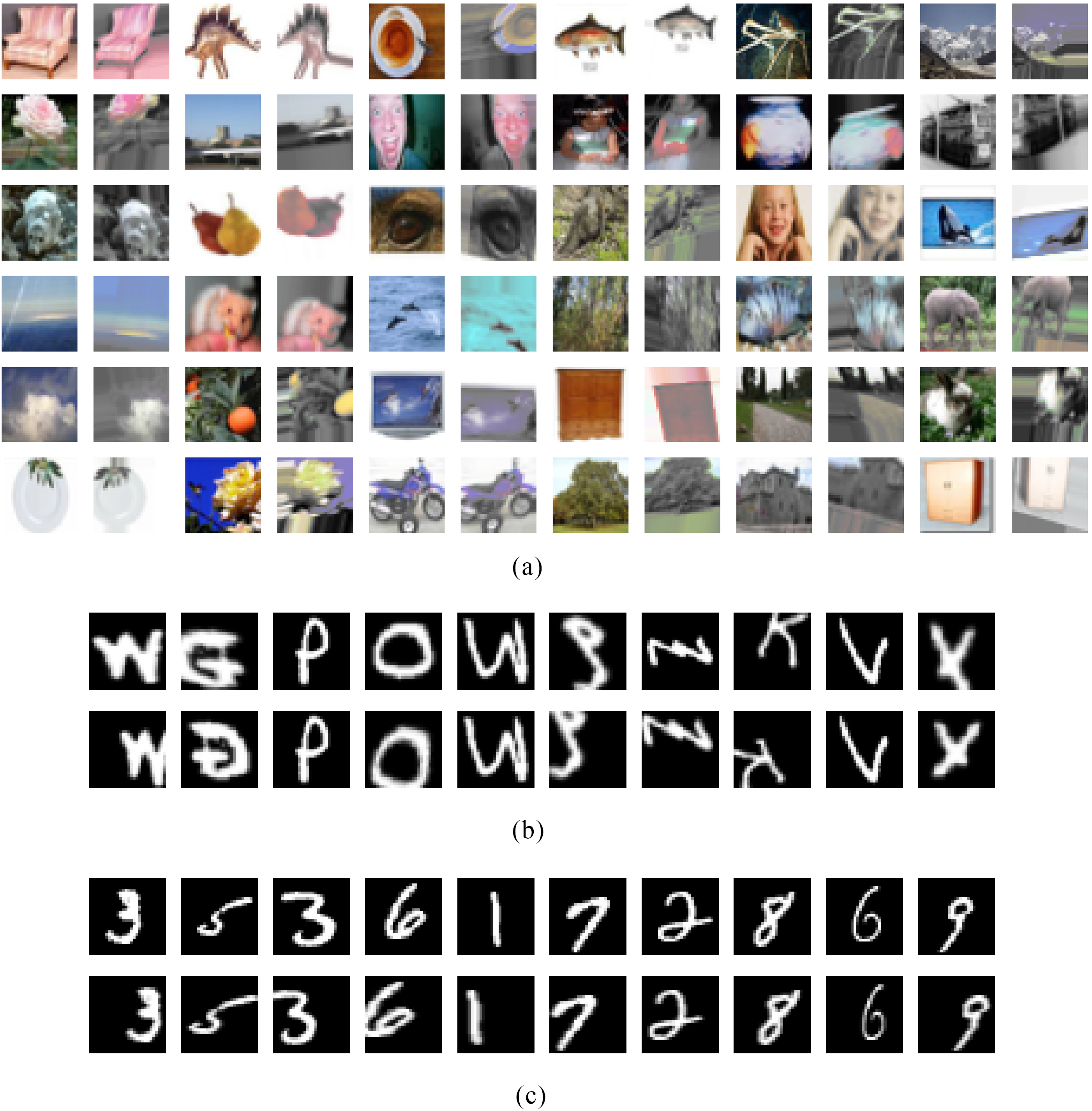}}
    
    \caption{Examples of pairs of CIFAR100/EMNIST/MNIST images, deformed using random affine transformations and color jitters. Notice that the transformations on MNIST is slight, but have already led to performance degradation on supervised networks.}
    \label{fig:deformation_all}%
\end{figure}

For the experiments with CIFAR \citep{krizhevsky2009learning} and EMNIST/MNIST \citep{cohen2017emnist, lecun-mnisthandwrittendigit-2010} we use random affine transformations to deform the images and get positive/negative pairs. The affine transformations include rotation, scaling, shifting, as well as flipping. For CIFAR100 in particular, we also apply random color jitters that changes the saturation and hue of the images. These deformations can be used to model the variety of views of a 3D object that form the `positive pairs' in reality. Some examples of the transformed CIFAR100 images can be found in Figure~\ref{fig:deformation_all} (a).

There are two types of transformations used in our experiments with EMNIST/MNIST. For creating the positive pairs in training the base encoder, we apply the same affine transformation we used for CIFAR100 to EMNIST, but without the color jitters. Then for testing the perturbation invariability of our SSL method, we apply random rotation, shifting and scaling to the MNIST dataset to creat a transformed MNIST dataset. The classifier is trained on the original MNIST and tested on the transformed MNIST. Examples of the transformations applied to EMNIST/MNIST can be found in Figure~\ref{fig:deformation_all} (b) and (c).

\section{More results with CIFAR}\label{sec:appendix-cifar}

\noindent {\bf Results with $\textsc{SimCLR}$.} We provide here the linear evaluation results with $\textsc{SimCLR}$ and biologically more plausible learning rules, to provide different levels of baselines for our experiments with the contrastive hinge loss. The results are obtained with exactly the same experimental settings for contrastive hinge loss. As can be seen from Table~\ref{tab:simclr} the results with $\textsc{SimCLR}$ are almost identical to what we obtained with contrastive hinge loss.

\begin{table} [h]
\small
\centering 
\begin{tabular}{c c c c} 
\toprule 
\textbf{Loss} & \textbf{Learning} & \textbf{Update} & \textbf{Classifier acc.}\\
\midrule 
$\textsc{SimCLR}$    & E2E & BP & 72.44\% \\
                &     & RF & 62.38\% \\
                &     & DTP & 71.00\% \\
                & GLL & BP/URF & 72.52\% \\
                &     & RF & 67.19\% \\
                & RLL & BP/URF & 68.99\% \\
                &     & RF & 66.39\% \\
\bottomrule 
\end{tabular}
\caption{Linear evaluation results with CIFAR10, using a $\textsc{SimCLR}$-trained base encoder with CIFAR100. We compare End-to-End (E2E), Greedy Layer-wise Learning (GLL) and Randomized Layer-wise Learning (RLL), and four updating methods: back-propagation (BP), Updated Random Feedbackau (URF), Random Feedback (RF) and Difference Target Propagation (DTP)} 
\label{tab:simclr} 
\end{table}

\noindent {\bf Effects of head size.} We run a set of preliminary experiments with end-to-end BP and the $\textsc{SimCLR}$ loss to investigate whether the head size in our network will affect the performance. The results in Table~\ref{tab:head} show that the size of the final projection head does not have a significant impact on linear evaluation performance. We thus keep the head size 64 for all our experiments:

\begin{table} [h]
\small
\centering 
\begin{tabular}{c c c c c} 
\toprule 
\textbf{Head Size} & 64 & 512 & 1024 & 2048\\
\midrule 
                & 69.87\% & 69.64\% & 69.97\% & 70.20\%\\

\bottomrule 
\end{tabular}
\caption{Investigation into head sizes. The linear evaluation results are obtained from a base encoder trained with CIFAR10 and classifier trained/tested on CIFAR10 as well. We use $\textsc{SimCLR}$  for these experiments.} 
\label{tab:head} 
\end{table}

\section{Code}

The code for our simulations and experiments can be found at:

\noindent
\href{https://github.com/C16Mftang/biological-SSL}{https://github.com/C16Mftang/biological-SSL}

\end{document}